\documentclass[10pt,twocolumn,letterpaper]{article}

\usepackage{iccv}
\usepackage{times}
\usepackage{epsfig}
\usepackage{graphicx}
\usepackage{amsmath}
\usepackage{amssymb}
\usepackage{multirow}
\usepackage{rotating}
\usepackage{booktabs}
\usepackage{subcaption}
\usepackage[font=small]{caption}
\usepackage{paralist}
\usepackage{enumitem}
\usepackage{verbatim}
\usepackage[normalem]{ulem}
\usepackage{dirtytalk}
\usepackage[toc,page]{appendix}

\usepackage{color}
\definecolor{orange}{rgb}{1,0.5,0}
\newcommand{\deshraj}{\textcolor{orange}}
\usepackage[pagebackref=true,breaklinks=true,letterpaper=true,colorlinks,bookmarks=false]{hyperref}

\iccvfinalcopy 

\def\iccvPaperID{242} 

\begin{document}

\title{It Takes Two to Tango: Towards Theory of AI's Mind}

\author{\textbf{Arjun Chandrasekaran}\thanks{\,\,\,Denotes equal contribution.}$\,\,^{,1}$ \qquad \textbf{Deshraj Yadav} $^{*,1,}$\thanks{\,\,\,Work done at Virginia Tech.} \qquad 
\textbf{Prithvijit Chattopadhyay}$^{*,1,}$\footnotemark[2] \\ 
\textbf{Viraj Prabhu}$^{*,1,}$\footnotemark[2] \qquad \textbf{Devi Parikh}$^{1,2}$ \vspace{0.005\textwidth} \\
$^1${\fontsize{11}{12}\selectfont Georgia Institute of Technology}
\qquad $^2${\fontsize{11}{12}\selectfont Facebook AI Research}
\\
{\tt\small \{carjun, deshraj, prithvijit3, virajp, parikh\}@gatech.edu}}


\maketitle

\begin{abstract}
Theory of Mind is the ability to attribute mental states (beliefs, intents, knowledge, perspectives, etc.) to others and recognize that these mental states may differ from one's own. 
Theory of Mind is critical to effective communication and to teams demonstrating higher collective performance. 
To effectively leverage the progress in Artificial Intelligence (AI) to make our lives more productive, it is important for humans and AI to work well together in a team. Traditionally, there has been much emphasis on research to make AI more accurate, and (to a lesser extent) on having it better understand human intentions, tendencies, beliefs, and contexts. The latter involves making AI more human-like and having it develop a theory of our minds.
In this work, we argue that for human-AI teams to be effective, humans must also develop a theory of AI's mind (ToAIM) -- get to know its strengths, weaknesses, beliefs, and quirks. 
We instantiate these ideas within the domain of 
Visual Question Answering (VQA). 
We find that using just a few examples (50), lay people can be trained to better predict responses and oncoming failures of a complex VQA model. 
We further evaluate the role existing explanation (or interpretability) modalities play in helping humans build ToAIM. Explainable AI has received considerable scientific and popular attention in recent times. 
Surprisingly, we find that having access to the model's internal states -- its confidence in its top-k predictions, explicit or implicit attention maps which highlight regions in the image (and words in the question) the model is looking at (and listening to) while answering a question about an image -- do not help people better predict its behavior.
\end{abstract}

\vspace{-20pt}
\section{Introduction}

\begin{figure}[t]
 \centering 
 \includegraphics[width=1\linewidth]{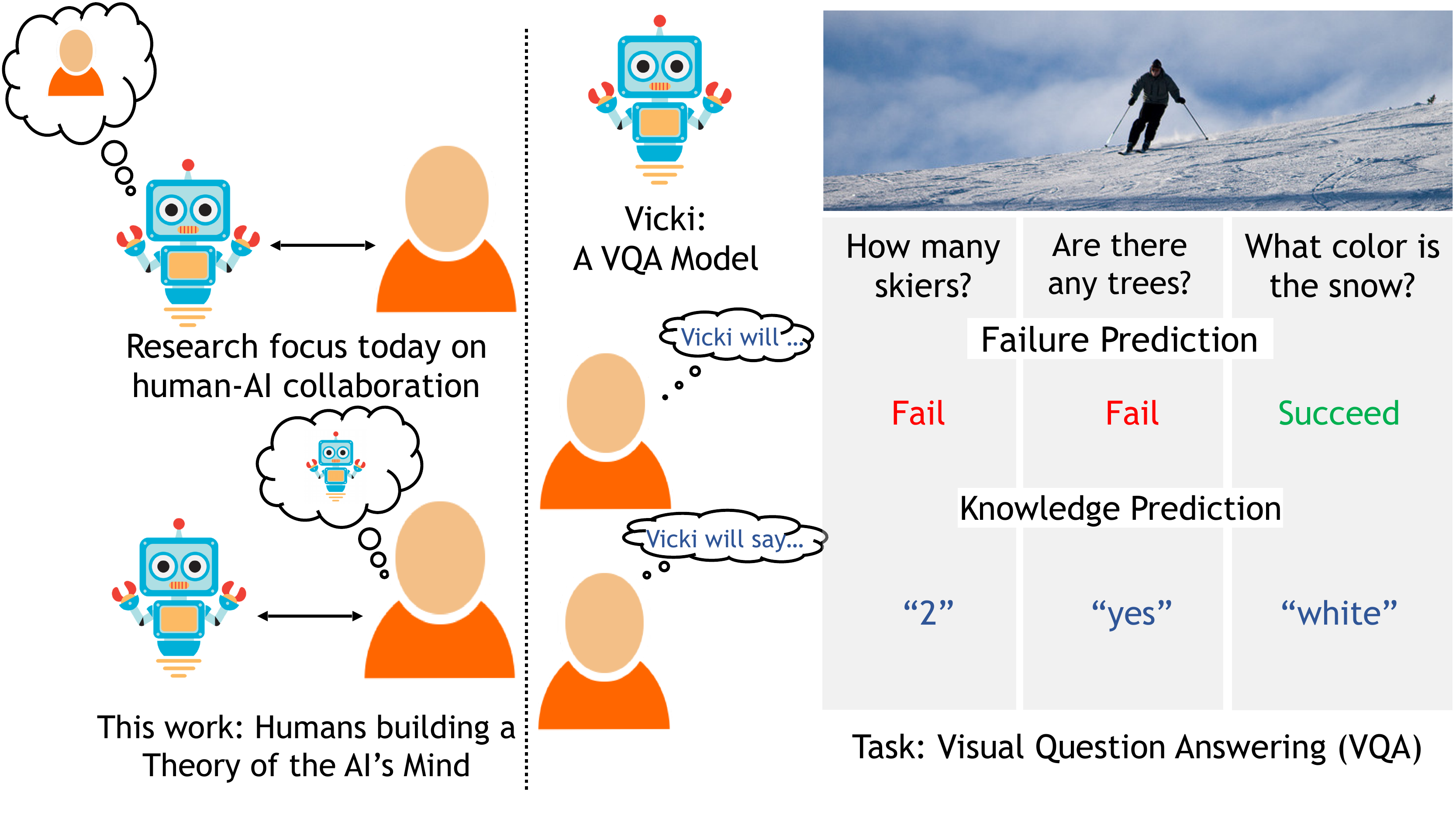}
 \caption{Current emphasis of research in human-AI collaboration is on AI modeling a human teammate's mental state (left top). We argue that for human-AI teams to be effective, humans must also have a model of the AI's strengths, weaknesses, and quirks. That is, humans must develop a Theory of AI's Mind (left bottom). In this paper, we instantiate these ideas in the context of Visual Question Answering (right). Human subjects predict the success or failure (Failure Prediction), and output responses (Knowledge Prediction) of a VQA model (we call Vicki). 
 }
\vspace{-16pt}
\label{fig:teaser}
\end{figure}

\noindent 
The capacity to attribute mental states\footnote{including beliefs, intents, desires, knowledge, etc.} to other agents that are different from one's own, is called \emph{Theory of Mind}~\cite{premack1978does}.
Effective communication involves considering a teammate's background knowledge, abilities, preferences and modifying one's interactions accordingly~\cite{grosz2012question}. 
Indeed, recent studies~\cite{engel2014reading, woolley2010evidence} conclude that the most effective teams are those with members who, among other traits, demonstrated good Theory of Mind abilities. 
\par
As AI progresses, we find ourselves working with AI agents increasingly often. 
Intelligent virtual assistants like 
Siri, 
Cortana, 
Google Assistant,  
and Alexa 
make our lives more convenient. Doctors collaborate with IBM's Watson~\cite{ferrucci2013watson,savvy2015watson}, dividing work based on their expertise to make better 
 informed diagnoses~\cite{bestdoctorsWatson}. Visually-impaired users are starting to rely on computer vision algorithms to interpret the world around them~\cite{wu2017automatic,microsoftCognitiveServices,bigham2010vizwiz}.
In-vehicle AI in autonomous cars leverage humans' experience to make decisions in unpredictable situations~\cite{nissanAutonomous}.
\par
\par 
Clearly, in each of these cases it is critical for the human to have a sense for what the AI is good at (vs not), or when the AI might fail and should not be trusted. The human-AI team will be more effective if the human collaborating with the AI agent has a deeper understanding of the AI agent's behavior. However, AI research has traditionally placed much of the burden on the AI to play its part in the team: to be more accurate~\cite{he2016deep,redmon2016you,szegedy2016inception,ren2015faster,liu2016ssd}, more human-like~\cite{andreas2016reasoning,kazemzadeh2014referitgame,mutlu2006storytelling,ferraro2016visual,agrawal2016sort,ray2016question,parikh2013implied,berg2012understanding,chandrasekaran2016we}, understand our intentions~\cite{oliver2000bayesian, vondrick2016predicting, vondrick2016anticipating}, beliefs~\cite{eysenbach2016mistaken}, tendencies~\cite{dubey2015makes}, contexts~\cite{recasens2015they}, and mental states~\cite{el2005real,el2004mind}. 
\par
In this work, we argue that for human-AI teams to be effective, Theory of Mind must go both ways. Humans must also understand the AI's beliefs, knowledge, and quirks (See Fig.~\ref{fig:teaser}).
To clarify, we do not claim that a layperson should understand internal workings of the AI; only that building an intuition for its behavior is important in a collaborative setting, particularly when the AI is imperfect. Today, Siri, Alexa, and image captioning systems are examples of imperfect AI systems in production, while more sensitive applications (AI-assisted driving, diagnosis) are on the horizon. We propose that humans thinking of these as collaborations with an AI teammate, and working toward building Theory of (AI's) Mind skills is beneficial from the standpoint of both effectiveness and safety (for when such imperfect systems fail). Such a skill is likely to thus be valuable to both designers of the AI and its users. 


\par
Research suggests that to understand new entities such as a robot, humans project existing preconceptions and social constructs upon them~\cite{fussell2008people}. 
However, as other recent works have shown~\cite{das2016human,goodfellow2014explaining}, the behavior of an AI agent is often quite different from that of a human -- sometimes in ways that are surprising. Thus, inferences based on existing social constructs or preconceptions may fail while estimating the behavior of AI agents.
\par \noindent

Of late, explainable AI has received considerable attention from both the scientific community
and popular media. 
Our work evaluates the usefulness of such explanation modalities in the specific setting of human-AI teams. Most prior work on interpretability has focused on demonstrating the role of such explanations in improving trust. To the best of our knowledge, this is the first evaluation that measures the extent to which these modalities allow a human to build a mental model for the AI and predict its behavior.

%
\par
We consider an agent trained to perform the multi-modal task of Visual Question Answering (VQA)~\cite{antol2015vqa,malinowski2014multi}.
 Given an image and a free-form open-ended natural language question about the image, the AI agent's task is to answer the question accurately. Call this agent -- a VQA model -- Vicki. VQA is applicable to scenarios where humans (e.g., visually impaired users, surveillance analysts, etc.) actively elicit information from visual data. It naturally lends itself to human-machine teams. 
The human teammates in our experiments are from Amazon Mechanical Turk 
(AMT). We consider two tasks that we believe demonstrate the degree to which a human understands their AI teammate Vicki -- Failure Prediction and Knowledge Prediction. In Failure Prediction (FP), we show AMT subjects an image and a question about the image, and ask them to estimate if Vicki will correctly answer the question. Their ability to estimate Vicki's success or failure in a scenario is a measure of how well they understand its strengths and weaknesses. In Knowledge Prediction (KP), subjects are asked to estimate Vicki's exact response. Making an accurate estimation of the response of the agent requires a deeper understanding of its behavior.
\par  
We study the extent to which humans can accurately estimate the behavior of Vicki. Then, we explicitly aid humans in developing a theory of Vicki's mind by (1) familiarizing them with Vicki's actual behavior during a training phase and (2) exposing them to Vicki's internal states via several existing `explanation' modalities. 
We evaluate if these explanation modalities  aid humans in accurately estimating Vicki's behavior (FP and KP).

\par 
\noindent
\par
While Theory of (human) Mind might appear to involve many complex mental states beyond beliefs and knowledge (which are clearly applicable to AI), in practice, it is measured  by a fairly simple test
-- ``reading the mind in the eyes''~\cite{baron2001reading}\footnote{Involves looking at a photo of a human's eyes and choosing one of two adjectives that better describes the person's mental state.}. In a similar vein, we propose the two tasks of FP and KP as simple yet effective techniques to measure a person's understanding and estimation of an AI agent's behavior, i.e., their Theory of an AI's mind (ToAIM)\footnote{On the subject of whether AI can have a mind at all, a number of philosophers suggest that it can. For instance, in `society of mind'~\cite{minsky1988society}, Minsky says that a mind simply emerges as a result of complex interactions between many smaller non-intelligent entities which he calls agents.}.

\par \noindent
\textbf{Contributions.} The contributions of this work are:
\setdefaultleftmargin{0pt}{}{}{}{}{}
\begin{compactenum}
\item We advocate a line of research to study the extent to which humans can build a Theory of AI's Mind (ToAIM) and develop approaches to aid the process.
\item As a specific instantiation of this, we consider the problem of VQA where the AI's task is to answer a free-form natural language question about an image.
\item We conduct large-scale human studies to measure the effectiveness of training, and of different explanation modalities, in helping humans accurately predict the successes, failures,and output responses of a VQA model on question--image pairs. To the best of our knowledge, this is the first evaluation that measures whether interpretability mechanisms do, in fact, allow humans to build a model of AI. Our human studies infrastructure will be made available.
\item Our key findings are that (1) humans are indeed capable of predicting successes, failures, and outputs of the VQA model better than chance. (2) explicitly training humans to familiarize themselves with the model by using just a few examples improves their performance (3) existing explanation modalities do not enhance human abilities at predicting the model's behavior. 
\end{compactenum}
\section{Related Work}

\noindent
\textbf{AI with a theory of (human) mind.} A number of works in AI attempt to develop agents with an understanding of human characteristics and behavior.
AI agents employing computer vision have been trained to predict the motivations~\cite{vondrick2016predicting}, intentions~\cite{oliver2000bayesian}, actions~\cite{vondrick2016anticipating}, tendencies~\cite{dubey2015makes}, contexts~\cite{recasens2015they}, etc., of humans. 
In addition, Scassellati~\cite{scassellati2002theory} examines theories that explain the development of Theory of Mind in children and their applicability to building robots with similar capabilities. More recently, in the domain of abstract scenes, Eysenbach et al.~\cite{eysenbach2016mistaken} address the problem of identifying incorrect beliefs in people. The ability to identify false beliefs~\cite{wimmer1983beliefs} in other agents is considered an important milestone in the development of Theory of Mind in an agent~\cite{baron1999evolution}. Unlike these works where AI agents ``understand'' humans, our work addresses the converse problem -- to have humans understand AI agents, their quirks, weaknesses, and beliefs. 

\par 
\noindent
\textbf{Explainable AI.} Recently, there has been a thrust in the direction of ``explainable'' AI agents in vision-related tasks. 
\par 
\noindent
\textbf{Introspection vs Justification:} Generating explanations for classification decisions has attracted considerable interest. Several works propose introspective explanations based on internal states of a decision process~\cite{zeiler2014visualizing, selvaraju2016grad, goyal2016towards, zhou2014object}, while others generate justifications consistent with model outputs~\cite{ribeiro2016should,hendricks2016generating,park2016attentive}. Riberio et al.~\cite{ribeiro2016should} explain the predictions of a classifier by learning an interpretable model locally around the prediction. Hendricks et al.~\cite{hendricks2016generating} develop a justification system that produces explanations consistent with visual recognition decisions.
\textbf{Natural language vs Visual explanations:} Prior art has assessed the usefulness of natural language explanations of model decisions in improving model trust~\cite{hendricks2016generating}. MacLeod et al.~\cite{macleodunderstanding} investigate the role of \emph{phrasing} of a model's confidence in blind and visually impaired persons' trust in image captioning models. Park et al.~\cite{park2016attentive} propose a pointing and justification model for VQA that can both justify predictions in natural language and also point to visual evidence. 
\textbf{Explicit vs Implicit attention:}  There is a line of work in designing models that explicitly attend to relevant parts of their input for vision tasks such as object recognition~\cite{ba-attention-2015,mnih2014recurrent}, image captioning~\cite{xu2015show, cho2015describing}, and VQA~\cite{lu2016hierarchical,yang2016stacked,xu2016ask}. In contrast, recent work by Zhou et al.~\cite{zhou2016learning} and Selvaraju et al.~\cite{selvaraju2016grad} expose implicit attention for predictions from CNN-based models as visual explanations.
\par \noindent 
Across these works, the focus is on making AI agents more transparent and capable of explaining their decisions in order to build trust. In our work, we explore the role explanation modalities play in improving a human's model of the AI, as measured by the human's accuracy at predicting the AI's success, failure, and output responses.
\par \noindent 
\textbf{Failure Prediction.} There exists prior art that deals with building models that predict failure modes of systems~\cite{bansal2014towards,zhang2014predicting}. Whereas these works employ statistical models to predict failure modes of a \emph{base system}, we evaluate the role a training phase as well as explanation modalities play when \emph{humans} perform the same task. In addition to predicting the success or failure of AI agents, we also train humans to more accurately predict the ``knowledge'', i.e., the actual output of an AI agent.
\par \noindent
\textbf{Humans adapting to technology.} A few works~\cite{wang2016learning,pelikan2016nao} observe human strategies while adapting to the limited capabilities of an AI agent in interactive language games. For instance, in a human-AI game of charades, humans modify strategies such as word selection, turn length, and prosody, to adapt to the robot's limited perceptive abilities. While both these works observe that humans dynamically adapt their behavior while interacting with an AI on a particular task, in our work we explicitly measure to what extent humans have formed an accurate model of the AI. We also evaluate the role that explanation/interpretability modalities play in helping humans build a more accurate model.

\section{Meet Vicki}
\label{sec:meet_vicki}
We instantiate the idea of humans building a Theory of AI's Mind in the VQA task. Our AI agent (that we call Vicki) is a VQA model trained to answer a free-form natural language question about an image. 
Concretely, we use the VQA model by Lu et al.~\cite{lu2016hierarchical}. It is a hierarchical coattention model that models the question at multiple levels of granularity (words, phrases, entire question) and at each level, has explicit attention mechanisms on the image (“where to look”) as well as the question (“which words and phrases to listen to”). Among the different variants introduced in~\cite{lu2016hierarchical}, we use the alternating co-attention model trained with VGG-19~\cite{simonyan2014very} as the CNN to derive image-representations. 
\par
Vicki was trained on the VQA dataset~\cite{antol2015vqa} train split containing 248349 QI pairs, and outputs one of a 1000 possible answers (most frequent in the train split). Its accuracy on the VQA dataset (test-standard) is 62.2\%\footnote{http://www.visualqa.org/roe.html} (human accuracy is 83.3\%), which was the state-of-the-art at publication~\cite{lu2016hierarchical} and is still competitive today. 
Moreover, Vicki's image and question attention maps provide access to its `internal states' while making a prediction. These maps highlight the regions of the image and words of the question that Vicki attends to.
This presents an opportunity to assess the role such explanation modalities can play in aiding humans better predict Vicki's behavior. Among the various settings explored in~\cite{lu2016hierarchical}, we use the \emph{question-level} image and question attention maps in our experiments. 
\par \noindent
\textbf{Vicki is Quirky.} There are several factors that contribute to Vicki being quirky, in a predictable fashion. Some of these quirks are well-known in VQA literature~\cite{agrawal2016analyzing}.
\textbf{Vision is not perfect:} Vicki, like most other vision models, has a limited capability to understand the image. Observing Vicki's behavior during its failures demonstrates its quirks. For instance, when the question asks the color of a small object in the scene, say a soda can, Vicki may simply respond with the most dominant color in the scene. This is clearly evident when we observe the distribution of Vicki's responses across a diverse set of images~\cite{agrawal2016analyzing}.
\textbf{Language is not perfect:} Vicki has a limited capability to understand free-form natural language. Vicki seems to converge on a predicted answer after listening to just half the question 49\% of the time \cite{agrawal2016analyzing}. So in many cases, it answers questions based only on the first few words of the question alone~\cite{agrawal2016analyzing}.
\textbf{Vicki cannot reason:} Vicki has no mechanism to leverage external knowledge and reason about common sense. Vicki is poor at compositionality -- it is unable to disentangle and recompose concepts seen in training to generalize to unseen test concepts \cite{agrawal2016analyzing}. Vicki does not have an explicit counting mechanism~\cite{ChattopadhyayVR16}. So it often defaults to the popular answer ``2'' for ``How many'' questions.
\textbf{Vicki cannot say much:} Since Vicki is a 1000-way classifier, it only has a fixed set of utterances. 
\textbf{Vicki answers every question:} Vicki was trained only on questions that were relevant to the image. Thus, Vicki does not know how to say ``That doesn't make sense.'' or ``There \emph{is} no woman in this image.'' when asked ``What color is the woman's shirt?'' on an image that does not contain a woman. Thus, when posed with a question that is irrelevant to the image, Vicki is forced to provide an answer from its limited vocabulary.  Interestingly, because Vicki is a deterministic function of the question and image, observing its response across QI-pairs often gives us a sense for what it might be basing its responses on.
\textbf{Vicki may ignore the image:}  Vicki picks up on the language priors that are inherent in the world which are easier to leverage than complicated visual signals. For example, when the question ``What color is the banana?'' is asked, Vicki often ignores the image and answers ``yellow''.
\textbf{Vicki is biased:} Vicki is very likely to answer ``yes'' to a yes/no question, and answer ``white' to a ``what color'' question due to biases inherent in the VQA dataset that it was trained on~\cite{goyal2016making}. \\
\begin{figure}[t]
 \centering 
 \includegraphics[width=1\linewidth]{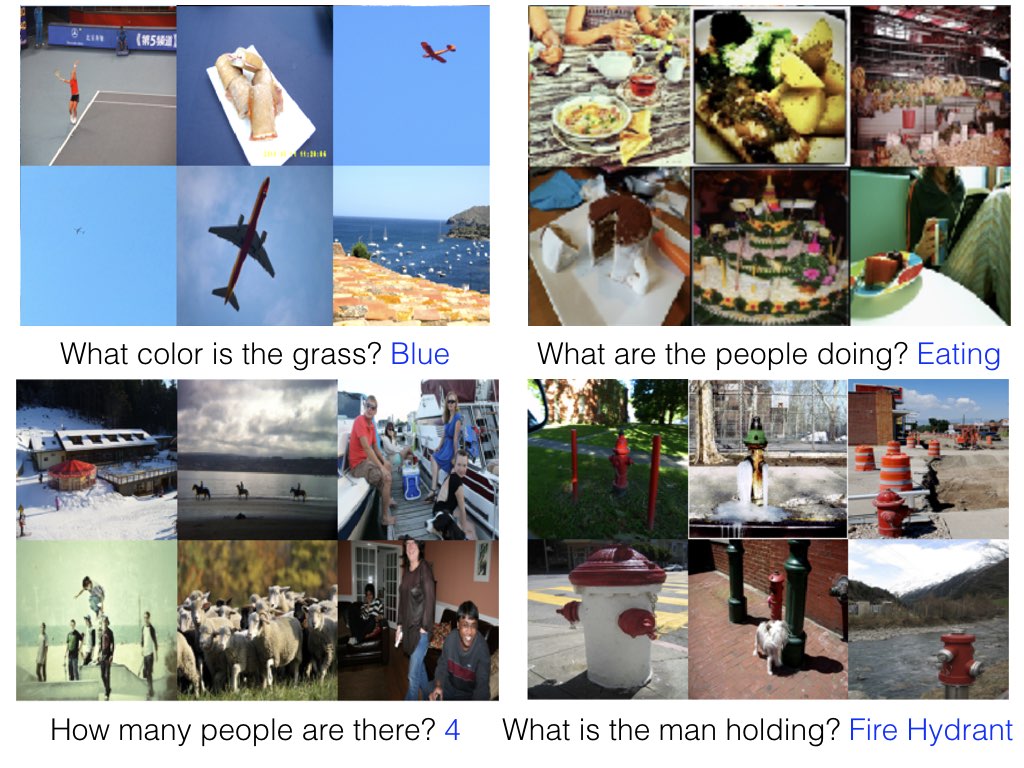}
 \vspace{-20pt}
 \caption{These montages highlight some of Vicki's quirks. For a given question, Vicki has the same response to each image in a montage. Common visual patterns (that Vicki presumably picks up on) within each montage are evident.}
 \vspace{-15pt}
\label{fig:montages}
\end{figure}
To get a sense for this, see Fig.~\ref{fig:montages}. 
The patterns are clear. In top-left, even when there is no grass, Vicki tends to latch on to one of the dominant colors in the image. For top-right, even when there are no people in the image, Vicki seems to respond with what people could \emph{plausibly} do in the scene if they were present. 
A priori, one (especially lay people) may not expect this. But when exposed to several examples of Vicki's responses, it is conceivable that subjects may begin to have an understanding of Vicki's behavior and consequently form a theory of its mind.
\section{Meet the tasks}
\label{sec:tasks}
%
\begin{figure*}[!t]
\setlength{\fboxsep}{0pt}
\setlength{\fboxrule}{0pt}
	\begin{subfigure}[t]{3.5in}	\fbox{\includegraphics[width=3.5in]{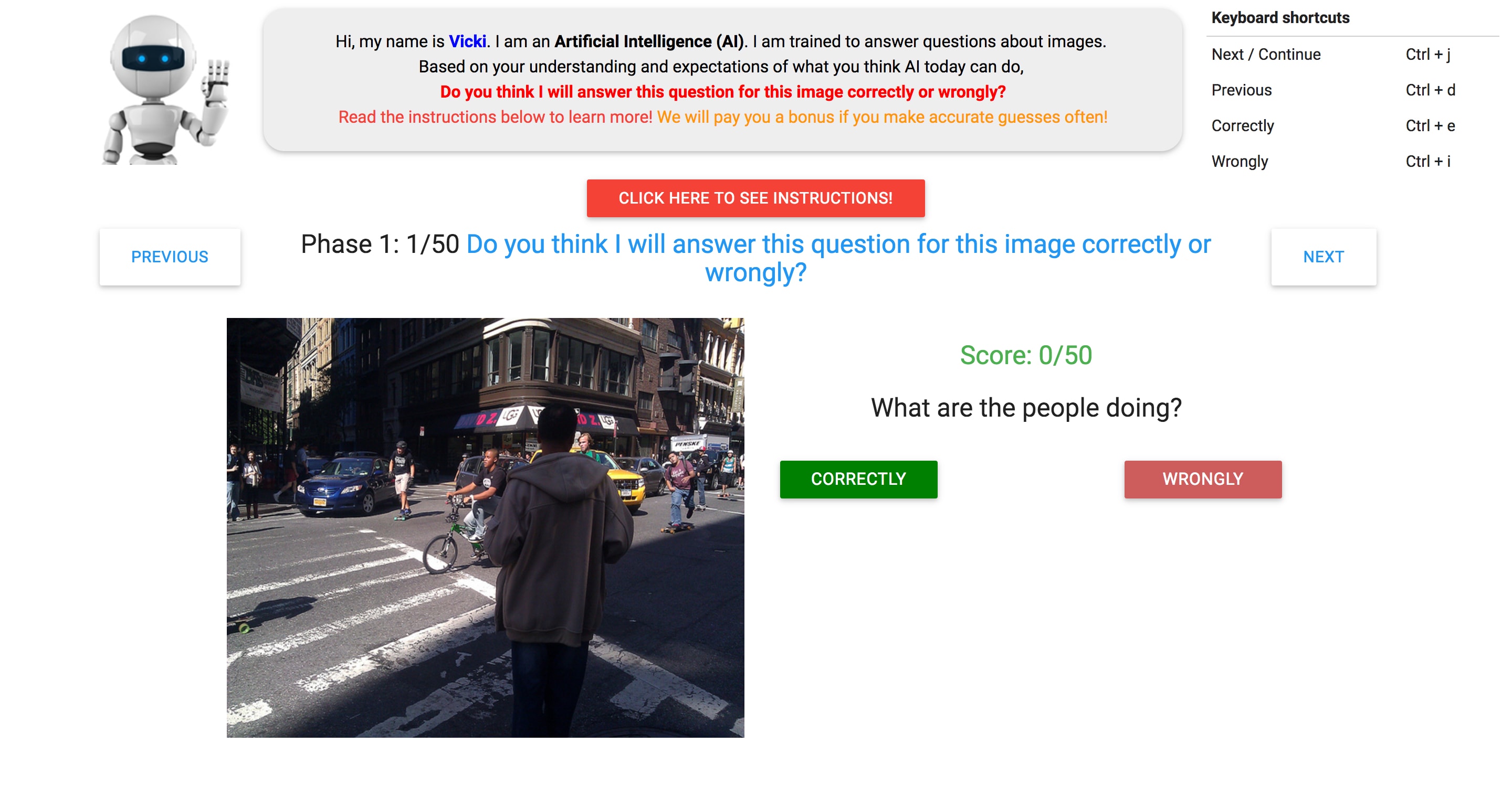}}
    \vspace{-15pt}
		\caption{The Failure Prediction interface.}
        \label{fig:fp_interface}
	\end{subfigure}
	\begin{subfigure}[t]{3.5in}
		\fbox{\includegraphics[width=3.5in]{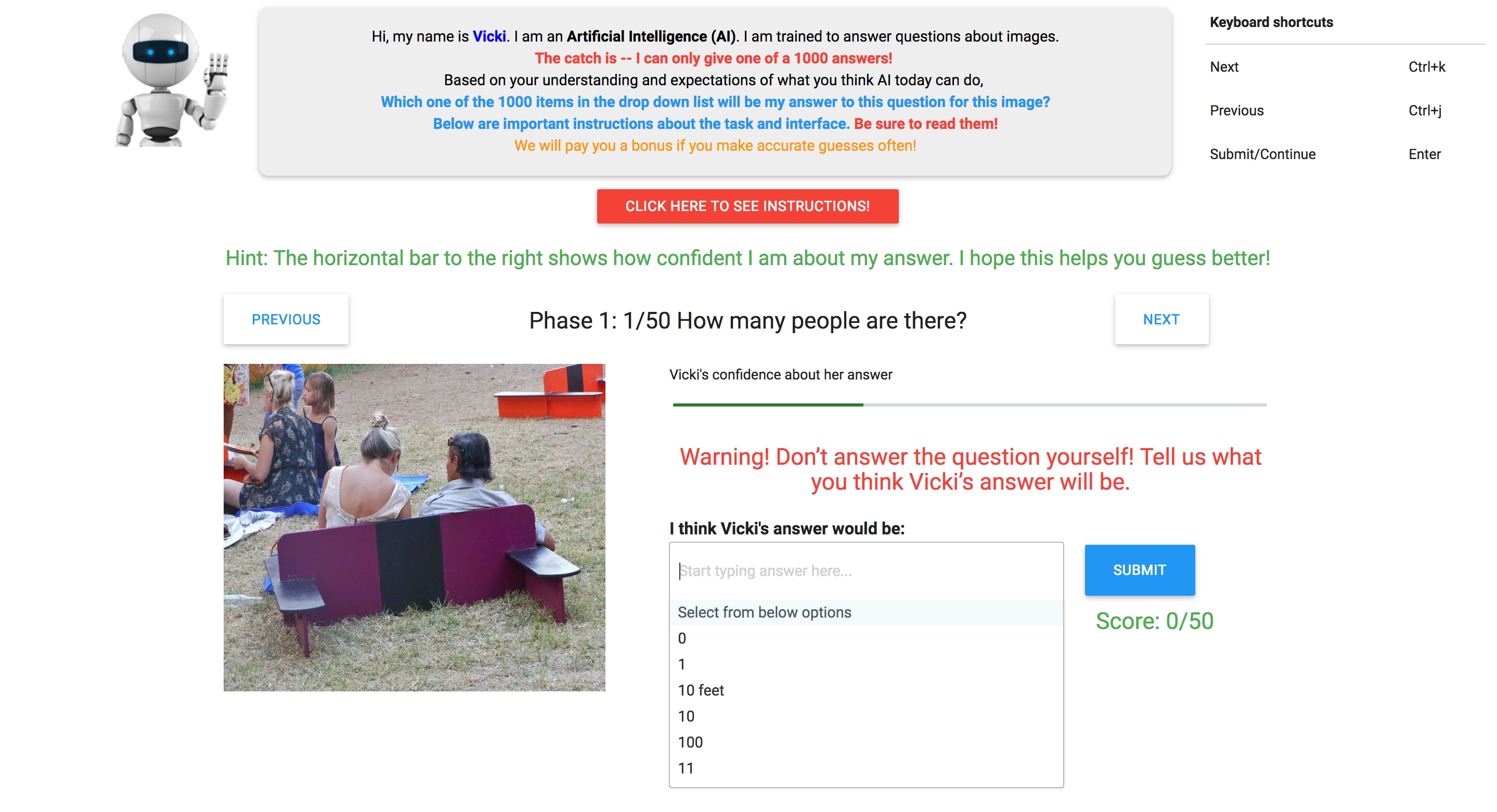}}
\vspace{-15pt}
		\caption{The Knowledge Prediction interface.}
        \label{fig:kp_interface}
	\end{subfigure}
    \vspace{-10pt}
\caption{(a) A person guesses if an AI agent (Vicki) will answer this question for this image correctly or wrongly. (b) A person guesses what Vicki's exact answer will be for this question for this image.
}
\vspace{-15pt}
\end{figure*}
We present two tasks that can measure a human's understanding of the capabilities of an AI agent such as Vicki. These tasks are especially relevant to human-AI teams since they are analogous to measuring if a human teammate's trust in an AI teammate is well-calibrated, and if a human can estimate the behavior of an AI in a specific scenario. 
\par \noindent
\textbf{Failure Prediction (FP).} In this task, we study the ability of a human to predict the success or failure of Vicki. That is, given an image and a question about the image, we measure how accurately a person can predict if Vicki will successfully answer the question.
A person can presumably predict the failure modes of Vicki reasonably well if they have a good sense of Vicki's strengths and weaknesses. A collaborator who performs well on this task can accurately determine whether they should trust Vicki's response to a question about an image. 
Please see a snapshot of the FP interface in Fig.~\ref{fig:fp_interface}. Note that we do not show the human what Vicki's predicted answer is.\footnote{Otherwise, given an image and question from the VQA dataset, it would be trivial for the human to verify if Vicki's predicted answer is right or wrong. See appendix for more details. 
}
\par 
\noindent
\textbf{Knowledge Prediction (KP).} In this task, we measure the capability of a human to develop a deeper understanding of Vicki's behavior. Given an image and a question, a person guesses Vicki's exact response (answer) from a set of its output labels (vocabulary). Recall that Vicki can only say one of a 1000 things in response to a question about an image. Please see a snapshot of the KP interface in Fig.~\ref{fig:kp_interface}. We provide subjects a convenient dropdown interface with autocomplete to choose an answer from Vicki's vocabulary of 1000 answers. 
\par 
In FP, a good understanding of Vicki's strengths and weaknesses might lead to good human performance. However, KP requires a deeper understanding of Vicki's behavior, rooted in its quirks and beliefs. In addition to reasoning about Vicki's failure modes, one has to guess its exact response for a given question about an image. Note that KP measures subjects' ability to take reality (the image the subject sees) and translate it to what Vicki might say. High performance at KP is likely to correlate to high performance at the reverse task -- take what Vicki says and translate it to what the image really contains. This can be very helpful when the visual content (image) is not directly available to the user. Explicitly measuring this is part of future work.
A person who performs well at KP has likely successfully modeled a more fine-grained behavior of Vicki than just modes of success or failure. In contrast to typical efforts where the goal is for AI to approximate human abilities, KP involves measuring a human's ability to approximate a neural network's behavior! 
\section{Perception of VQA}
To set the baseline, we  measure 
people's current estimates about VQA models. To this end, we briefly introduce Vicki to subjects as an ``AI trained to answer questions about images''. We then ask subjects to use their current understanding and expectation of what AI agents can do, to estimate the behavior of Vicki. 
We study the ability of humans to estimate Vicki's behavior via the FP and KP tasks. For both tasks, we randomly sample questions from the set of $\sim$1,400  most frequent questions in the validation set of the VQA dataset~\cite{antol2015vqa}. A description of our experimental setup for each task follows.

\par 
\subsection{Failure Prediction (FP)}
\label{subsec:fp}In this task\deshraj{,} we show subjects a question and an image on which this question was asked in the VQA dataset, and ask if they think Vicki's response to the question--image pair (QI--pair) would be correct or wrong. To get ground truth, similar to the VQA accuracy metric~\cite{antol2015vqa}, we check if Vicki's response matched at least 3 of 10 human-provided answers in the VQA dataset. 
Overall, a total of 88 unique subjects participated in our study, providing responses on 1000 QI--pairs. 
On average, subjects accurately guessed whether Vicki would answer the question correctly (success) or not (failure) 59.88\% of the time. The accuracy of always guessing success is 61.52\%. While subjects' performance seems lower than this, normalizing for the prior of each class (success vs. failure), always guessing success drops to 50\% but humans are at 54.24\%. 
This shows that even without prior exposure to Vicki, human subjects can predict its failure better than chance. 

\par
\begin{figure}[t]
 \centering 
 \includegraphics[width=1\linewidth]{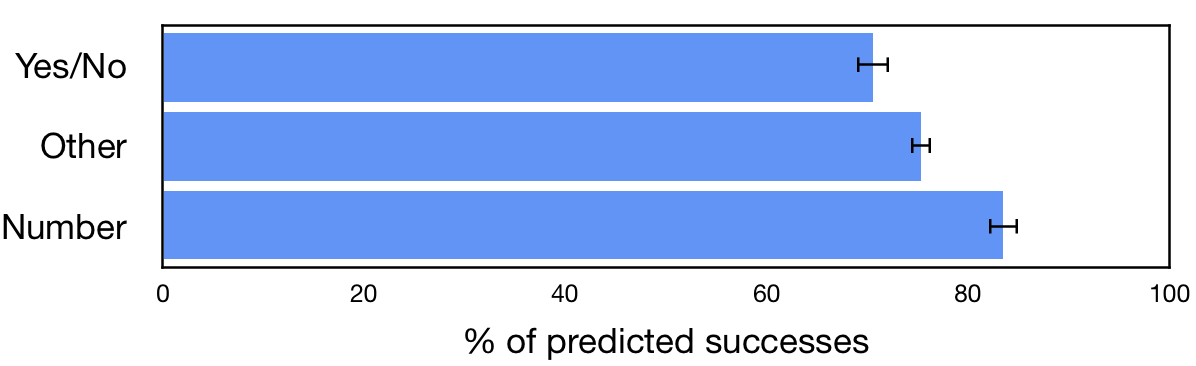}
 \vspace{-15pt}
 \caption{Optimism regarding Vicki: people's estimate of the AI's success in answering a question, across different answer-types.}
 \vspace{-15pt}
\label{fig:vqa_optimism}
\end{figure}
We further measure people's optimism about Vicki's abilities. Fig.~\ref{fig:vqa_optimism} shows the percentage of QI-pairs that subjects predicted Vicki would answer correctly for different answer types. We find that subjects expect Vicki to answer questions whose answers are numbers (e.g., counting questions) correctly quite often. Interestingly, today's VQA models are in fact quite ineffective at counting. The VQA leaderboard shows significant improvements in performance on ``other'' questions over time, but improvements on ``number'' questions has stalled. 
Overall, subjects demonstrated an average optimism -- as measured by \% of ``correctly'' (success) predictions -- of 75.46\%. 
%
%
\subsection{Knowledge Prediction (KP)}
\label{subsec:kp}
In the KP task, we ask subjects what they think Vicki would say in response to a question about an image. Note that the VQA dataset only contains questions about an image that are relevant to the image, as 
annotators were looking at the image while asking questions. So a question ``What color is the man's shirt?'' would only be asked for an image that contains a man wearing a shirt.

As an interesting twist intended to elicit Vicki's quirky behavior described in Sec.~\ref{sec:meet_vicki}, we also paired images with random (and likely irrelevant ~\cite{ray2016question}) questions (e.g., ``What are the people doing?'' on an image that may not contain people). Recall that Vicki is forced to respond with a limited vocabulary (one of 1000 answers). These samples are useful to measure a person's understanding of an agent's responses to any given stimulus -- including those that come from a distribution under which the agent has not been trained. Note that FP cannot be evaluated on irrelevant images. The notion of a ``correct'' answer is ill-defined if a question is not relevant to an image.

We performed the KP task\footnote{To control for familiarity, we ensured that subjects who perform a KP task are not allowed to perform an FP task.}
on 1000 QI--pairs (700 relevant and 300 irrelevant). We collected 25 responses to each pair. A total of 173 unique subjects participated in our study. The accuracy achieved by predicting Vicki's most popular answer (`yes') is 15.79\%. We found that subjects were able to accurately predict Vicki's response 24.81\% of the time.


\section{Familiarizing people with Vicki}
In this section we describe our experimental setup to familiarize subjects with Vicki's behavior. We approach this in two ways -- by providing instant feedback about Vicki's actual behavior on each QI pair once the subject responds, and by exposing subjects to various explanation modalities that reveal Vicki's internal states. 

\noindent
\textbf{Challenges.} Collecting data for this setup is challenging for a couple of reasons: (1) Each subject has to go through a training phase to become familiar with Vicki before we can test them. This results in each task on AMT being unusually long and expensive. It also reduces the subject pool down to those willing to participate in long tasks. (2) Once a subject does one task for us, they cannot do another task because the training / exposure to Vicki would leak over. This means we need as many subjects as tasks. This makes data collection quite slow. 
In light of these challenges, to systematically evaluate the roles of training and exposure to Vicki's internal states, we focus on a small set of questions.

\par \noindent
\textbf{Data.} We identify a subset of questions in the VQA~\cite{antol2015vqa} validation split that occur more than 100 times. We select 7 diverse questions from this subset that are representative of the different types of questions (counting, yes/no, color, scene layout, activity, etc.) in the dataset\footnote{What kind of animal is this? What time is it? What are the people doing? Is it raining? What room is this? How many people are there? What color is the umbrella?}. 
For each of the 7 questions, we then sample a set of 100 images.
For FP, the 100 images per question are random samples from the set of images on which the question was asked in the VQA validation split (VQA-val). For the KP task, these 100 images are random images from VQA-val. Ray et al.~\cite{ray2016question} found that randomly pairing an image with a question in the VQA dataset results in about 79\% of pairs being irrelevant.
Recall that this combination of relevant and irrelevant question-image pairs allows us to test subjects' ability to develop a robust understanding of Vicki's behavior across a wide variety of inputs.

\par \noindent 
\textbf{Task setup.} Each human study is comprised of 100 QI-pairs where a single question is asked across 100 images. The motivation behind keeping the question constant is to make it easier for the subject to pick up trends in Vicki's responses across images. The annotation task is broken down into a train phase where the person is shown 50 QI-pairs, and a test phase where we evaluate subject's performance on the remaining 50 QI-pairs. 
\par \noindent
\subsection{Does feedback help?} 
\label{subsec:feedback} 
To familiarize subjects with Vicki, we provide them with instant feedback during the train phase. Immediately after a subject responds to a QI--pair, we show them whether Vicki actually answered the question correctly or not (in FP) or what Vicki's response was (in KP). In the train phase, subjects are also shown a live score of how well they are doing and are allowed to scroll through feedback for previous images (of course, they are not allowed to change their answers to previous images). Once training is complete, no further feedback (including running score) is provided and subjects are asked to draw from the intuition they have built in training to best answer all questions in the test phase. Subjects are also paid a bonus if they do particularly well. 


To evaluate the role of instant feedback, we have 2 subjects do our study with and without instant feedback each, for each of the questions (7) and each task (FP and KP). This results in a total of 28 human studies (with 28 unique human subjects). Even without feedback, subjects still go through all 100 images. 

\begin{figure}[t]
 \centering 
 \includegraphics[width=1\linewidth]{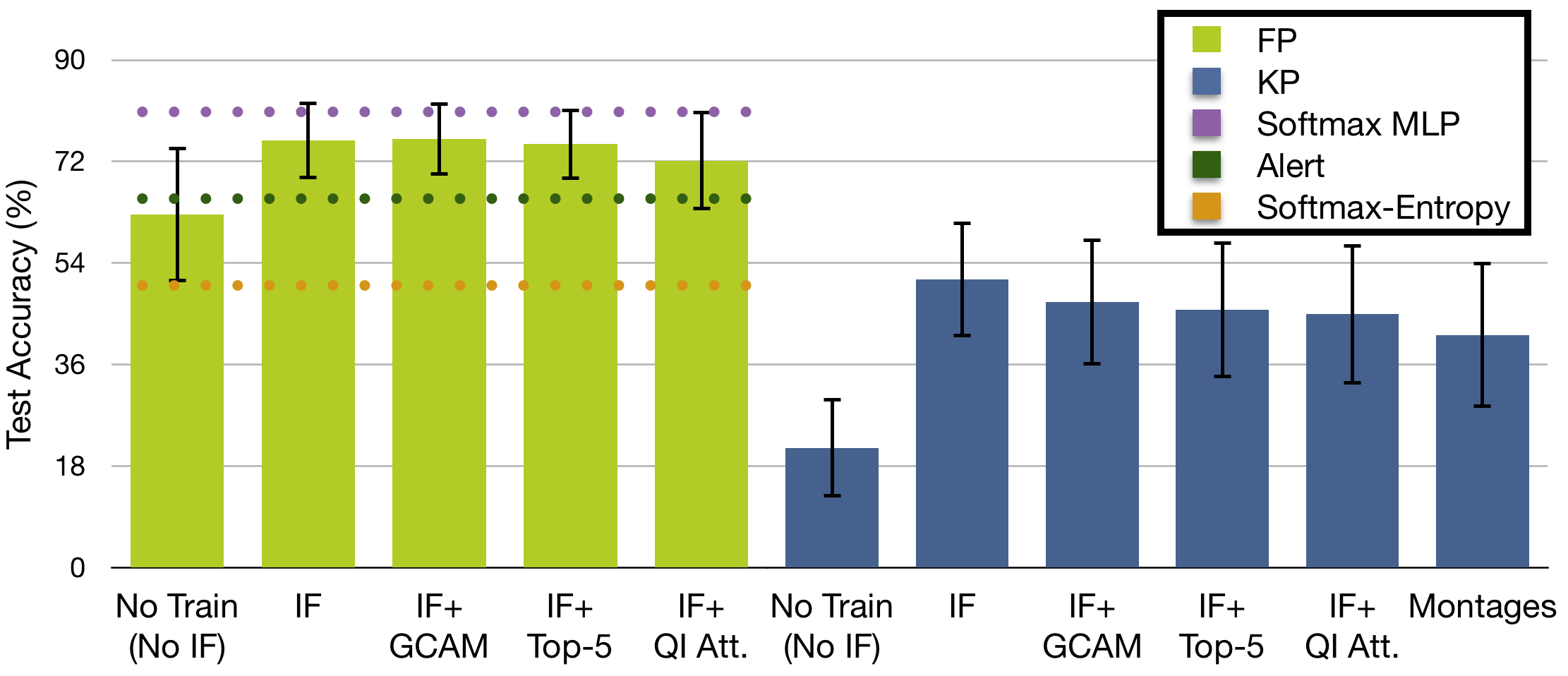}
 \vspace{-20pt}
 \caption{Average performance across subjects for both tasks: Failure Prediction (FP) and Knowledge Prediction (KP), with and without instant feedback (IF), and with various explanation modalities. Error bars are 95\% confidence intervals from 1000 bootstrap samples. Note that the dotted lines are various machine approaches applied to FP. 
 }
 \vspace{-15pt}
\label{fig:xai}
\end{figure}

In FP, always answering ``correctly'' would result in an accuracy of 58.29\%. We find that subjects do slightly better and achieve 62.66\% accuracy on FP, even without prior familiarity with Vicki (No Train).  Thus, subjects are already slightly better calibrated with an AI's capabilities than unbridled optimism (or pessimism). Further, we find that subjects that receive training as instant feedback (IF) achieve 13.09\% (absolute) higher mean accuracies than those who do not (see Fig~\ref{fig:xai}); IF vs No Train (No IF) for FP (green). \par

In KP, answering each question with Vicki's most popular answer overall (`no') would lead to an accuracy of 13.4\%.
Additionally, answering each question with Vicki's most popular answer \emph{for that question}\footnote{Vicki's most frequent answer (in the train set) to each question is as follows: What kind of animal is this? (Dog). What time is it? (Daytime). What are the people doing? (Standing). Is it raining? (No). What room is this? (Kitchen). How many people are there? (1). What color is the umbrella? (Black). } leads to an accuracy of 31.43\%. Interestingly, subjects who are unfamiliar with Vicki (No Train) achieve 21.27\% accuracy -- better than the most popular answer overall prior, but worse than the question-specific prior over Vicki's answers. The latter is understandable as subjects unfamiliar with Vicki do not know which of its 1000 possible answers are more likely a priori for each question.

We find that mean absolute performance in KP with IF is 51.11\%, 29.84\% higher than KP without IF (see Fig~\ref{fig:xai}; IF vs No Train (No IF) for KP (blue)). Subjects thus considerably outperform both the `most popular answer' and `most popular answer per question' priors. It is apparent that just from a few (50) training examples, subjects learn to generalize beyond Vicki's favorites among it's vocabulary of 1000 answers. Additionally, the 29.84\% improvement over No Train for KP is significantly larger than that for FP (13.09\%). This is understandable because a priori (No Train setting), KP is a much harder task as compared to FP, due to the increased space of possible subject responses given a QI-pair, and the combination of relevant and irrelevant QI-pairs in the test phase. 


Questions such as `Is it raining?' have strong language priors -- to these Vicki often defaults to the most popular answer (`no'), irrespective of image. We observe that on such questions, subjects perform considerably better in KP once they develop a sense for Vicki's inherent bias via instant feedback. For open-ended questions like `What time is it?', feedback helps subjects (1) narrow down the 1000 potential options to the subset that Vicki typically answers with  -- in this case time periods such as `daytime' rather than actual clock times and (2) identify correlations between visual patterns and Vicki's answer (as seen in Fig.~\ref{fig:montages}). In other cases like `How many people are in the image?' the space of possible answers is clear a priori, but after IF subjects realize Vicki is not good at detailed counting but does base its count predictions on coarse signals of the scene layout.

\par
In Sec.~\ref{sec:meet_vicki}, we described how montages (refer to ~Fig.\ref{fig:montages}) help highlight Vicki's quirks. In order to test the effectiveness of such montages as a teaching tool, we also experimented with a modification of the KP + IF setting (two unique subjects per question participated in this setting, resulting in an additional 14 human studies). In the train phase of this new setting, instead of individual images, subjects are shown a series of \emph{montages}, each containing 4 to 16 images across which Vicki gave the \emph{same} answer to the question. The objective remains the same -- to guess what that answer was (with IF provided after each guess). The test phase is kept identical to the KP + IF test phase, with a single image per question and no IF. We find that subjects achieve 41.6\% mean accuracy in the test phase of this setting, which is lower than the mean accuracy in the test phase of the KP + IF setting (51.1\%). Interestingly, mean accuracy in the train phase of the montage setting is 68.7\%, significantly higher than the mean accuracy in the train phase of the KP + IF setting (49.3\%). This seems to indicate that while montages make it much easier to guess Vicki's response correctly by picking out patterns (as seen in Fig.~\ref{fig:montages_1} and Fig.~\ref{fig:montages_2}), the focus on identifying commonalities between groups of images interferes with the ability to pick up on image-level patterns. As a result, subjects do not generalize well to  individual images at test time, resulting in worse performance. Keeping the train and test tasks identical (individual images in both cases) is more effective.

\noindent 
\textbf{VQA Researchers.} Just as an anecdotal point of reference, we also conducted experiments across experts with varying degree of familiarity with agents like Vicki. We observed that a VQA researcher had an accuracy of 80\% versus a computer vision (but not VQA) researcher who had 60\% in a shorter version of the FP task without instant feedback. Clearly, familiarity with Vicki plays a critical role in how well a human can predict its oncoming failures or successes.
\subsection{Do explanation modalities help?} 
\label{subsec:x_modalities}

In this section, we briefly describe the different explanation modalities that we utilize to expose Vicki's internal states to the human subject. 
In addition to an image and question about the image, we also show the subject one of the three explanation modalities described below. 
Subjects are asked to use these as hints to perform the task (FP or KP) more accurately, and can leverage the training phase (with instant feedback and a running score) to learn how to best do so.

\par 
\begin{figure}[t]
 \centering 
 \includegraphics[width=1\linewidth]{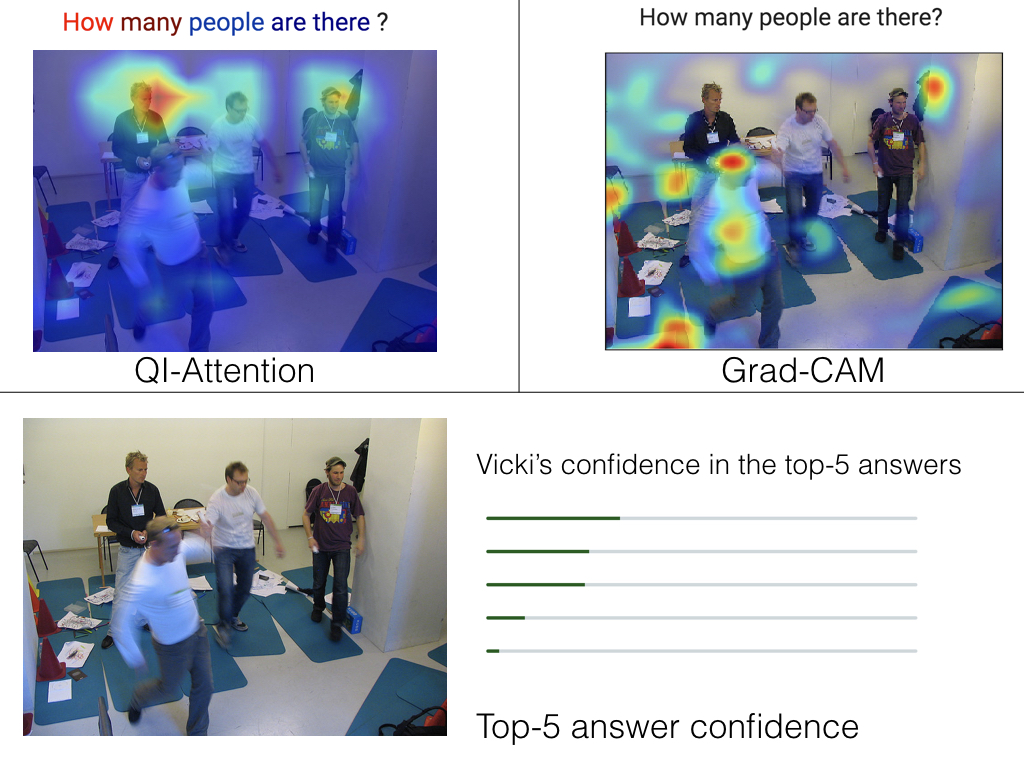}
 \vspace{-20pt}
 \caption{Screenshots of the interfaces of different explanation modalities that we show subjects.}
 \vspace{-15pt}
\label{fig:xai_modal}
\end{figure}
We experiment with 3 qualitatively different explanation modalities (see Fig~\ref{fig:xai_modal}): 
\par \noindent
\textbf{Confidence of top-5 predictions.} We show subjects Vicki's confidence in its top-5 answer predictions from its vocabulary as a bar plot\footnote{Of course, we don't show the actual top-5 predictions, just the confidence in the predictions.}. If Vicki is relatively more confident in its top-1 prediction, it is more likely to be right. If Vicki is confused about the top-5 predictions, it is more likely to be wrong.  
\textbf{Attention maps.} Recall that Vicki is the co-attention VQA model proposed by Lu et al~\cite{lu2016hierarchical} which jointly reasons about image and question attention (Sec~\ref{sec:meet_vicki}). Thus, along with the image we show subjects the spatial attention map over the image that indicates the regions that Vicki is looking at and an attention map over each word of the question highlighting the relative importance of words in the question for Vicki, while producing an answer. 
We show subjects a legend to interpret what the colors in each attention map indicate.
\textbf{Grad-CAM.} In contrast to explicit attention maps described above, we experiment with an implicit attention map. We use the CNN visualization technique by Selvaraju et al.~\cite{selvaraju2016grad}
, using the attention maps corresponding to Vicki's most confident answer.

We have 2 subjects perform each of our tasks (2) for each of the explanation modalities (3) for each question (7) resulting in a total of 84 tasks (and unique subjects). Across all studies (including those described in earlier sections), we have collected over $65k$ responses from 415 unique subjects. Conducting studies in-house in controlled environments at this scale would be prohibitive.


To put human FP accuracies (using explanation modalities) in perspective, we experiment with a few automatic approaches to detect Vicki's failure from its internal states. We find that a decision stump on Vicki's confidence in its top answer or the entropy of its 1000-way softmax output results in FP accuracy of 60\% on our test set. We train a Multilayer Perceptron (MLP) neural network on Vicki's output softmax and predict success vs failure. This achieves an FP accuracy of 81\%\footnote{Showing a visualization of this score to a human may make for a good ``explanation modality'' for FP! Exploring this is part of future work.}. Training an MLP which takes as input question features (average word2vec embeddings~\cite{mikolov2013distributed} of words in the question) concatenated with image features (fc7 from VGG-19) to predict success vs failure (which we call \text{ALERT} following ~\cite{zhang2014predicting}) achieves an FP accuracy of 65\%. Note that these methods are trained on about 66\% of the VQA--val set ($\sim$81k examples, rest used for validation). Human subjects are trained on only 50 examples. We only report machine results to put human accuracies in perspective. We do not draw any inferences about the relative capabilities of both

Accuracies of subjects in the test phase of both tasks (FP and KP) for different settings of the explanation modalities are summarized in Fig.~\ref{fig:xai}. Recall, all studies that include an explanation modality also include instant feedback\footnote{In real--world settings, we consider familiarizing via instant--feedback, followed by showing explanation modalities, as the natural progression for acquainting subjects with Vicki. Hence, we evaluate the role explanation modalities play on top of instant feedback. Nevertheless, for sake of completeness, studying the effect of showing explanation modalities on subject performance, independent of instant feedback, is part of future work.} (IF) and a running score during training. For reference, we also show performance of subjects with no explanation modality both with and without IF. We observe that on both tasks, subjects shown explanation modalities along with IF show no statistically significant improvement in performance over those shown just IF. In fact, in some cases performance is worse. While piloting these tasks ourselves, we found that it was easy to ``overfit'' to the explanation modalities and hallucinate patterns when none may exist. 
While the works introducing some of these modalities assessed their interpretability qualitatively or measured their role in improving human trust, our preliminary hypothesis is that these modalities may not yet help human-AI teams be more accurate in a goal-driven collaborative setting because they do not yet help humans predict the AI's behavior more accurately.
\section{Conclusion}
We posit that as AI makes progress, human-AI teams are imperative. We argue that for these teams to be effective, it is not only important for the AI to be capable of modeling the intentions, beliefs, strengths and weaknesses of the human, but \emph{also} for the human to build a Theory of the AI's Mind (ToAIM). \textbf{Take-home message \#1}: We should pursue research directions to help humans build models of the strengths, weaknesses, quirks, and tendencies of AI. 
We instantiate these ideas in the domain of Visual Question Answering (VQA). We propose two tasks that help measure the extent to which a human ``understands'' a VQA model (we call Vicki) -- Failure Prediction (FP) and Knowledge Prediction (KP) -- where given an input instance (question--image pair) the human has to predict whether Vicki will answer the question correctly or not, and what Vicki's exact answer will be. We evaluate the roles that familiarity with Vicki and explanation modalities that expose the internal states of Vicki play. \textbf{Take-home message \#2}: Lay people indeed get better at predicting Vicki's behavior using just a few (50) ``training'' examples. \textbf{Take-home message \#3}: Surprisingly, existing explanation modalities that are popular in computer vision do not help make Vicki's failures more predictable. In fact, humans seem to overfit to the additional information provided and perform slightly worse at KP in the presence of explanation modalities. \textbf{Take-home message \#4}: Clearly, much work remains to be done in developing improved explanation modalities that do in fact help make AI more predictable to a human.

As AI improves, does a user's ToAIM become outdated? In informal studies, we found human FP/KP performance with one VQA model to generalize to another. Further, we envision “ToAIM building” being a continual exercise integrated in the deployment of AI, with every major update involving a familiarization stage.

This work just scratches the surface, and numerous avenues of further exploration remain. Studying other vision models (AI agents in general) at varying points on the interpretability vs performance spectrum for other tasks, evaluating other existing explanation modalities, and conducting human studies at an even larger scale are natural extensions. Relevant to the increased interest in building interpretable models, this work presents novel opportunities to evaluate explanation modalities grounded in specific tasks (FP and KP). Finally, it would be exciting to close the loop and evaluate the extent to which improved human performance at FP and KP translates to improved success of human-AI teams at accomplishing a shared goal. Co-operative human-AI games may be a natural fit for such evaluation.

\par \noindent
\textbf{Acknowledgements.} We thank Satwik Kottur for his help with data analysis, and for many fruitful discussions. We thank Aishwarya Agrawal and Akrit Mohapatra for their help with experiments. We would also like to acknowledge the workers on Amazon Mechanical Turk for their effort. This work was funded in part by an NSF CAREER award, ONR YIP award, Sloan Fellowship, ARO YIP award, Allen Distinguished Investigator award from the Paul G. Allen Family Foundation, Google Faculty Research Award, Amazon Academic Research Award to DP. The views and conclusions contained herein are those of the authors and should not be interpreted as necessarily representing the official policies or endorsements, either expressed or implied, of the U.S. Government, or any sponsor.


{\footnotesize
\bibliographystyle{ieee}
\bibliography{toaim}
}
\appendix
\section*{Appendix}

\section{Introduction}
This supplementary material is organized as follows: We first discuss various visual recognition scenarios in which a human might rely on an AI, and motivate the need for Theory of AI's Mind (ToAIM) in those scenarios. Next, we include video demonstrations of our FP and KP interfaces. We then provide more qualitative examples of montages (introduced in Fig.2 of main paper) that highlight Vicki's quirks, and additionally share insights on Vicki from subjects who completed the tasks. Finally, we describe an AMT survey we conducted to gauge public perception of AI, and provide a list of questions and analysis of results.
\section{Visual Recognition Scenarios and Applicability of ToAIM}
In general, one might wonder why a human would need Vicki to answer questions if they are already looking at the image. This may be true for the VQA dataset, but outside of that there are scenarios where the human either does not know the answer to a question of interest (e.g., the species of a bird), or the amount of visual data is so large (e.g., long surveillance videos) that it would be prohibitively cumbersome for them to sift through it. Note that even in this scenario where the human does not know the answer to the question, a human who understands Vicki's failure modes from past experience would know when to trust its decision. For instance, if the bird is occluded, or the scene is cluttered, or the lighting is bad, or the bird pose is odd, Vicki will fail. Moreover, the idea of humans predicting the AI's failure (and ToAIM in general) also applies to other scenarios where the human may not be looking at the image, and hence needs to work with Vicki (e.g., blind user, or a human working with a tele-operated robot). In these cases too, it would be useful for the human to have a sense for the contexts and environments and/or kinds of questions for which Vicki can be trusted. In this work, as a first step, we focus on the first scenario where the human is looking at the image and a question while predicting Vicki's failures and responses.
\section{Interfaces}


To enable readers to experience the FP and KP tasks firsthand, we provide a link to the interfaces we used to train subjects: \url{https://deshraj.github.io/TOAIM/}. We also provide links to videos demonstrating each task: FP -- \url{https://youtu.be/Dcs7GOmTAns} and KP -- \url{https://youtu.be/f_1ikwCuG4Q}. Note that for illustration, we show just a single setting of the respective task in each interface and video.

\section{Vicki's Quirks}
\begin{figure*}[t]
 \centering 
 \includegraphics[width=\textwidth,height=\textheight,keepaspectratio]{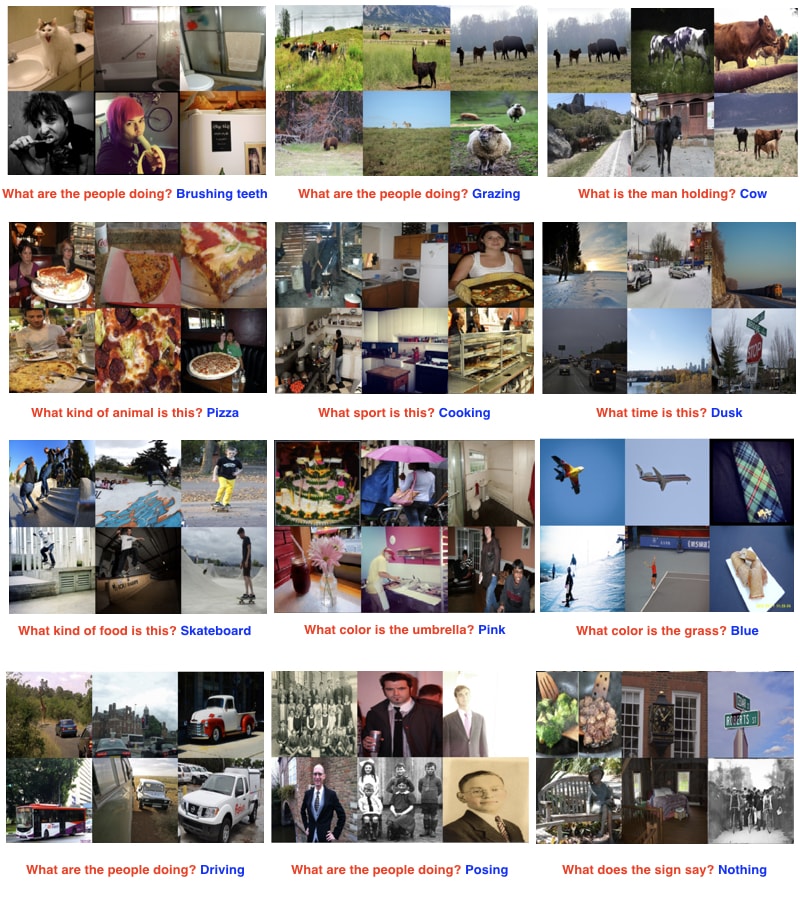}
 \caption{Given a question (red) we show images for which Vicki gave the same answer (blue) to the question to observe Vicki's quirks.  
 }
 \vspace{30pt}
\label{fig:montages_1}
\end{figure*}

\begin{figure*}[t]
 \centering 
 \includegraphics[width=\textwidth]{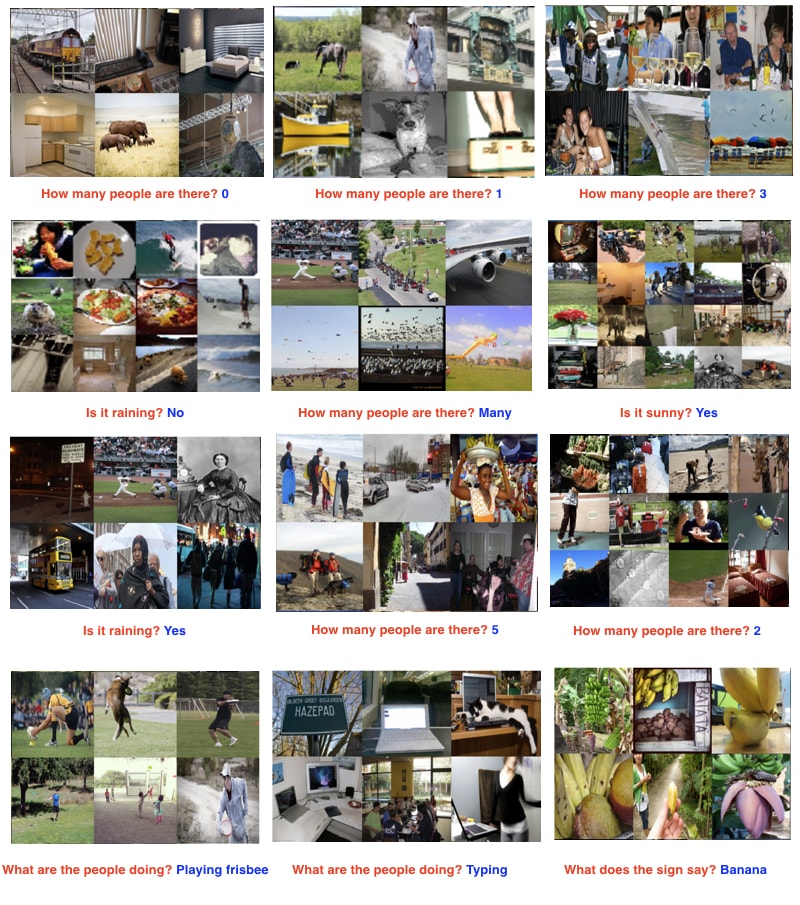}
 \caption{Given a question (red) we show images for which Vicki gave the same answer (blue) to the question to observe Vicki's quirks. 
 }
 \vspace{30pt}
\label{fig:montages_2}
\end{figure*}

We present more examples in Fig.~\ref{fig:montages_1} and Fig.~\ref{fig:montages_2} that highlight Vicki's quirks. Recall that there are several factors which lead to Vicki being quirky, many of which are well known in VQA literature~\cite{agrawal2016analyzing}. As we can see across both examples, Vicki exhibits these quirks in a somewhat predictable fashion. At first glance, the primary factors that seem to decide Vicki's response to a question given an image are the properties and activities associated with the salient objects in the image, in combination with the language and the phrasing of the question being asked. This is evident when we look across the images (see Fig.~\ref{fig:montages_1} and~\ref{fig:montages_2}) for question-answer (QA) pairs such as -- \emph{What are the people doing? Grazing}, \emph{What is the man holding? Cow} and \emph{Is it raining? No}. As a specific example, notice the images for the QA pair \emph{What color is the grass? Blue} (see Fig.~\ref{fig:montages_1}) -- Vicki's response to this question is the most dominant color in the scene across all images even though there is no grass present in any of them. Similarly, for the QA pair \emph{What does the sign say? Banana} (see Fig.~\ref{fig:montages_2}) -- Vicki's answer is the salient object across all the scenes.
\par
Interestingly, some subjects did try and pick up on some of the quirks and beliefs described previously, and formed a mental model of Vicki while completing the Failure Prediction or Knowledge Prediction tasks. We asked subjects to leave comments after completing a task and some of them shared their views on Vicki's behavior. We share some of those comments below. 

\par \noindent
The abbreviations used are Failure Prediction (FP), Knowledge Prediction (KP) and Instant Feedback (IF).
\newpage
\begin{compactenum}
\item \textbf{FP}
\begin{itemize}
\item \emph{\say{These images were all pretty easy to see what animal it was. I would imagine the robot would be able to get 90\% of the animals correct, unless there were multiple animals in the same photo.}}
\item \emph{\say{I think the brighter the color the more likely they are to get it right. Multi-colored, not so sure.}}
\item \emph{\say{I'd love to know the answers to these myself.}}
\end{itemize}
\item \textbf{FP + IF}
\begin{itemize}
\item \emph{\say{This is fun, but kind of hard to tell what the hints mean. Can she determine the color differences in multi-colored umbrellas or are they automatically marked wrong because she only chooses one color instead of all of the colors? It seems to me that she just goes for the brightest color in the pic. This is very interesting. Thank you! :)}}
\item \emph{\say{I didn't quite grasp what the AI's algorithm was for determining right or wrong. I want to say that it was if the AI could see the face of the animal then it guessed correctly, but I'm really not sure.}}
\end{itemize}
\item \textbf{FP + IF + Explanation Modalities}
\begin{itemize}
\item \emph{\say{Even though Vicki is looking at the right spot doesn't always mean she will guess correctly. To me there was no rhyme or reason to guessing correctly. Thank you.}}
\item \emph{\say{I think she can accurately know a small number of people but cannot know a huge grouping yet.}}
\item \emph{\say{I would be more interested to find out how Vickis metrics work. What I was assuming is just color phase and distance might not be accurate.}}
\end{itemize}
\item \textbf{KP}
\begin{itemize}
\item \emph{\say{Time questions are tricky because all Vicki can do is round to the nearest number.}}
\item \emph{\say{there were a few that seemed like it was missing obvious answers - like bus and bus stop but not bus station.  Also words like lobby seemed to be missing.}}
\end{itemize}
\item \textbf{KP + IF}
\begin{itemize}
\item \emph{\say{Interesting, though it seems Vicki has a lot more learning to do. Thank you!}}
\item \emph{\say{This HIT was interesting, but a bit hard. Thank you for the opportunity to work this.}}
\end{itemize}
\item \textbf{KP + IF + Explanation Modalities}
\begin{itemize}
\item \emph{\say{You need to eliminate the nuances of night time and daytime from the computer and choose one phrasing "night" or "day" Vicki understands. The nuance keeps me and I'm sure others obtaining a higher score here on this task.}}
\item \emph{\say{I felt that Vickie was mistaken as to what some colors were for the first test which probably carried over and I tried my best to recreate her responses.}}
\end{itemize}
\item \textbf{KP + IF + Montages}
\begin{itemize}
\item \emph{\say{I am not sure that I ever completely understood how Vicki thought. It seemed it had more to do with what was in the pictures instead of the time of day it looked in the pictures. If there was food, she chose noon or morning, even though at times it was clearly breakfast food and she labeled it noon.}}
\item \emph{\say{It doesn't seem very accurate as I made sure to count and took my time assessing the pictures.}}
\item \emph{\say{it is hard to figure out what they are looking for since there isn't many umbrellas in the pictures}}
\end{itemize}
\end{compactenum}

On a high-level reading through all comments, we found that subjects felt that Vicki's response often revolves around the most salient object in the image, that Vicki is bad at counting,
 and that Vicki often responds with the most dominant color in the image when asked a color question. In Fig.~\ref{fig:cmt_wcloud}, we show a word cloud of all the comments left by the subjects after completing the tasks. From the comments, we observed that subjects were very enthusiastic to familiarize themselves with Vicki, and found the process engaging. Many thought that the scenarios presented to them were interesting and fun, despite being hard. We used some basic elements of gamification, such as performance-based reward and narrative, to make our tasks more engaging; we think the positive response indicates the possibility of making such human-familiarization with AI engaging even in real-world settings. 

\section{Perception of AI}
\label{sec:public_perception}
Before introducing people to Vicki and gauging their expectations from a modern VQA system, we attempted to assess their general impressions of present-day AI. 

\begin{figure*}[t!]
\centering
\setlength{\fboxsep}{0pt}
\setlength{\fboxrule}{0pt}
\begin{subfigure}[t]{3.45in}
  	\fbox{\includegraphics[width=3.45in, height=2.25in]{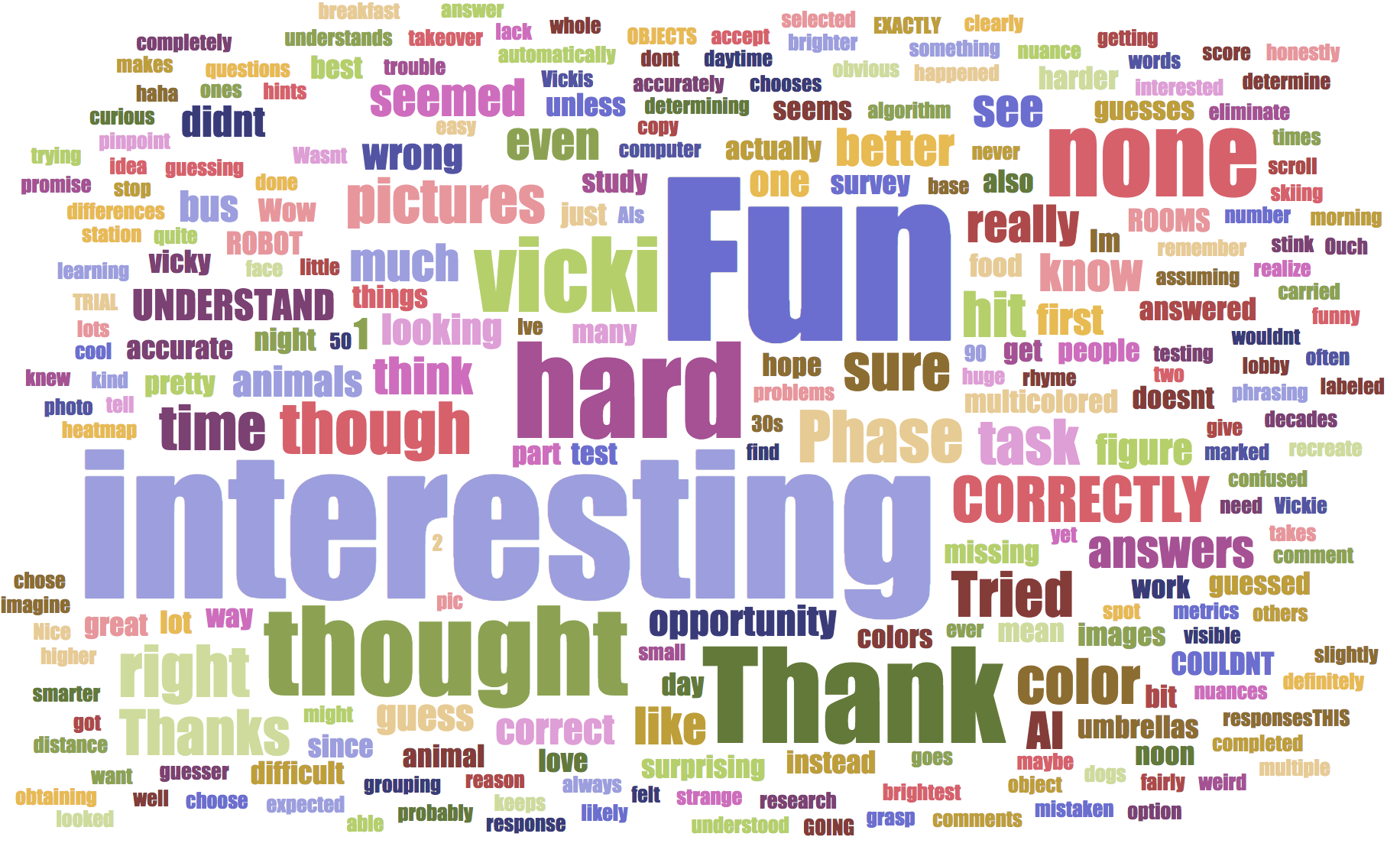}}
	\caption{We show a word cloud of all the comments left by  subjects after completing the tasks across all settings. From the frequency of positive comments about the tasks, it appears that subjects were enthusiastic to familiarize themselves with Vicki.}		
	\label{fig:cmt_wcloud}
\end{subfigure}	
\hspace{0.1in}    
\begin{subfigure}[t]{3.2in}
	\fbox{\includegraphics[width=3.2in, height=2.25in]{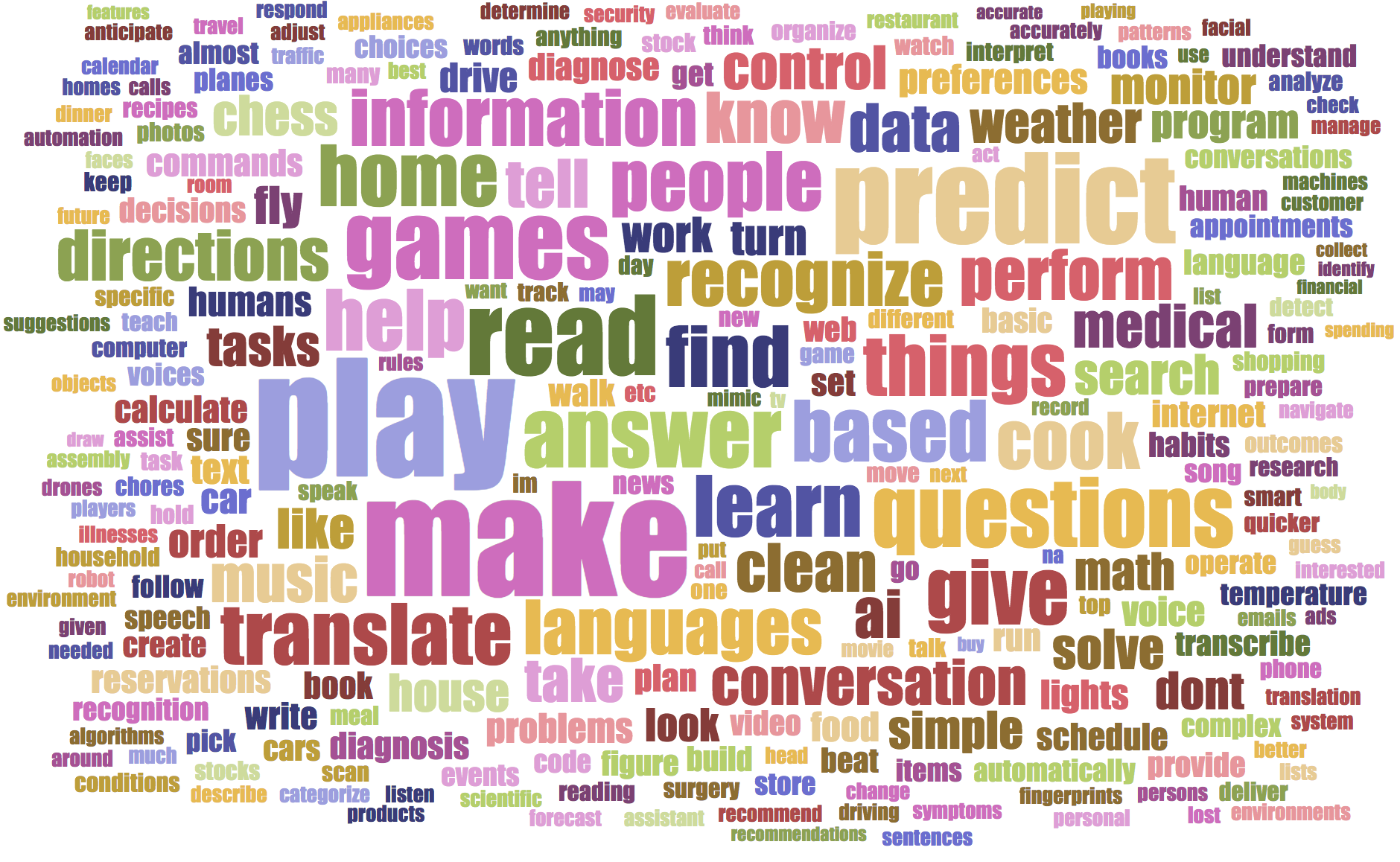}}
    \caption{A word cloud of subject responses to ``Name three things that you think AI today can do.''}
    \label{fig:ai_now}
\end{subfigure}
\vspace{0.05in} 
\begin{subfigure}[t]{3.2in}
	\fbox{\includegraphics[width=3.2in, height=2.25in]{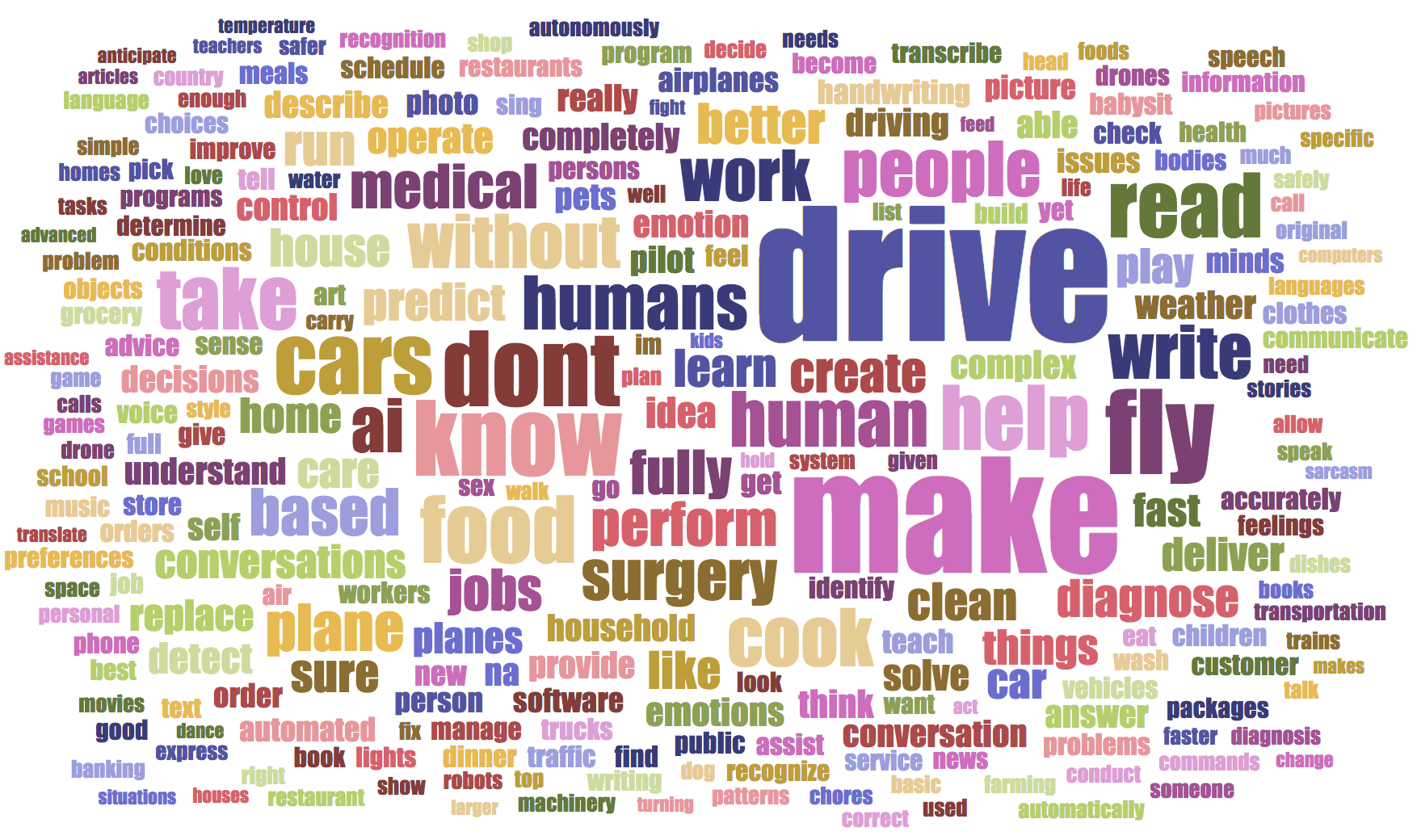}}
    \caption{A word cloud of subject responses to ``Name three things that you think AI today can't yet do but will be able to do in 3 years.''}
    \label{fig:ai_3_yrs}
\end{subfigure}         
\hspace{0.1in}    
\begin{subfigure}[t]{3.2in}
	\fbox{\includegraphics[width=3.2in, height=2.25in]{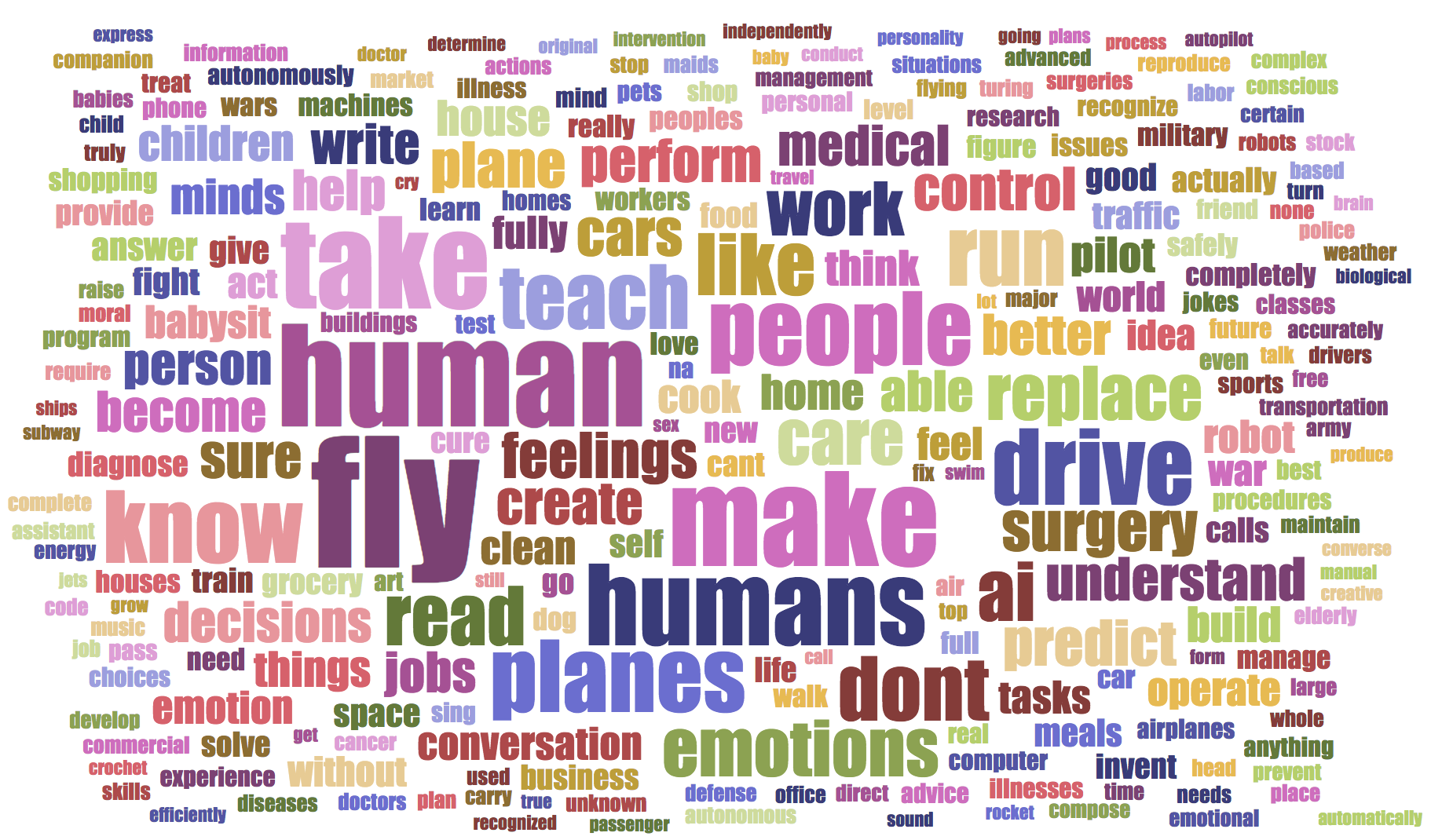}}
    \caption{A word cloud of subject responses to ``Name three things that you think AI today can't yet do and will take a while ($>$ 10 years) before it can do it.''}
	\label{fig:ai_10_yrs}
\end{subfigure}        
\vspace{0.05in}    
\begin{subfigure}[t]{3.2in}
	\fbox{\includegraphics[width=3.2in, height=2.25in]{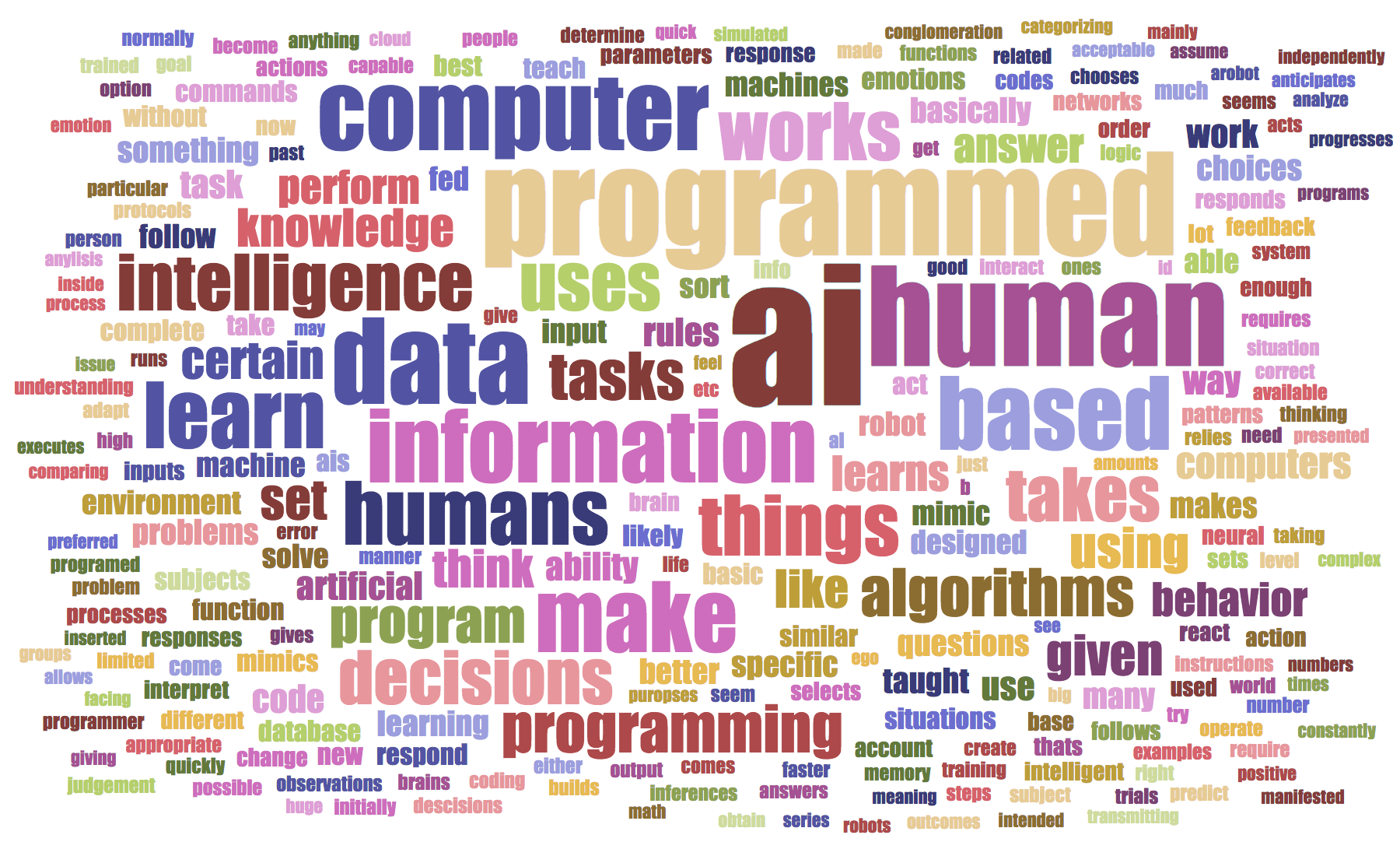}}
    \caption{A word cloud of subject responses when asked to describe how \emph{they} think AI today works. }
	\label{fig:how_ai_works}
\end{subfigure}
\caption{Word clouds corresponding to responses from humans for different questions.}
\label{fig:perception_of_ai}
\end{figure*}

We asked each subject to fill out a survey with questions aimed to collect three types of information:

\begin{compactenum}
\item \textbf{Background information.} We collected basic demographic information such as age, gender, educational qualifications, type of residential area, and profession. We also collected socio-economic background information such as employment status and income group. 
\item \textbf{Familiarity with computers and AI.} We asked subjects if their jobs involved computers, if they knew how to program, how much time they spent in front of a computer or smartphone, and their familiarity with popular AI assistants such as \emph{Siri}, \emph{Alexa} and \emph{Google Assistant}. We also asked if they were aware of recent advances in AI, especially those trending in popular media, such as \emph{Watson}~\cite{ferrucci2012introduction}, \emph{AlphaGo}~\cite{silver2016mastering}, machine learning, and deep learning.
\item \textbf{Estimates of AI's capabilities.} We asked subjects their duration and source of exposure to AI and gathered their impressions on the capabilities of modern-day AI systems on a range of tasks. We also asked them about their understanding, trust and sentiment towards modern AI systems, as well as their expectations and predictions for AI in the future.
\end{compactenum}

\begin{figure*}[t]
 \centering 
 \includegraphics[width=1\textwidth]{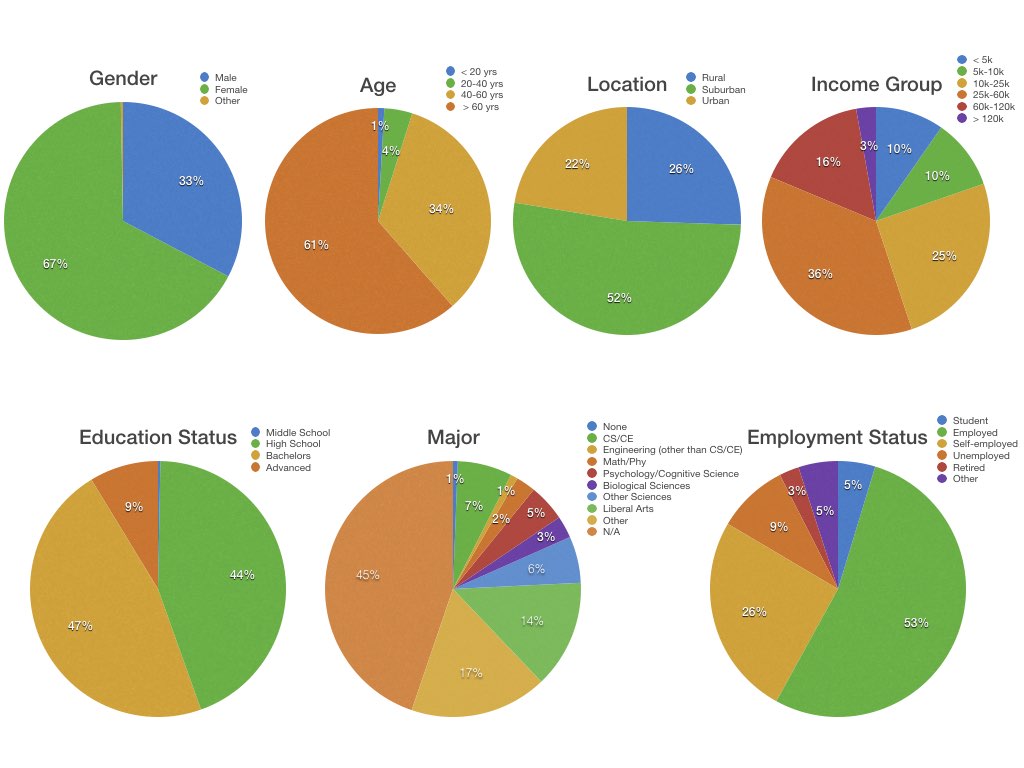}
 \vspace{-20pt}
 \caption{Population Demographics (across 321 subjects)}
 \vspace{-15pt}
\label{fig:pop_demog}
\end{figure*}

\begin{figure*}[t]
 \centering 
 \includegraphics[width=1\textwidth]{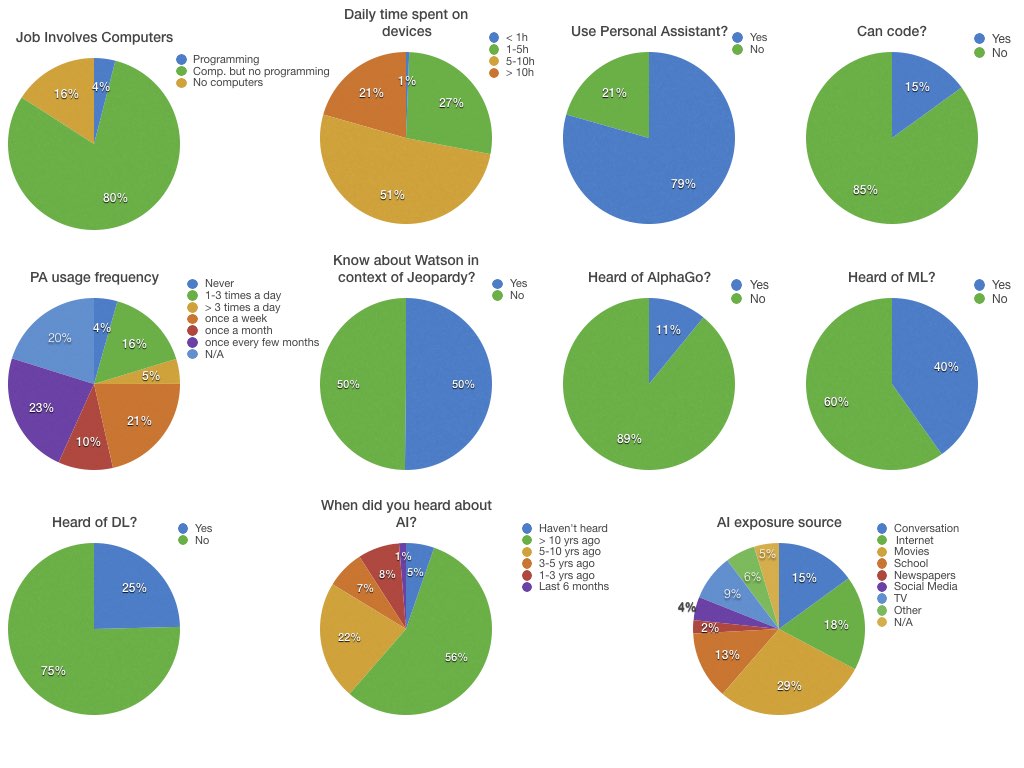}
 \vspace{-20pt}
 \caption{Technology and AI exposure (across 321 subjects)}	
 \vspace{-15pt}
\label{fig:tech_expo}
\end{figure*}

\begin{figure*}
 \centering 
 \includegraphics[width=1\textwidth]{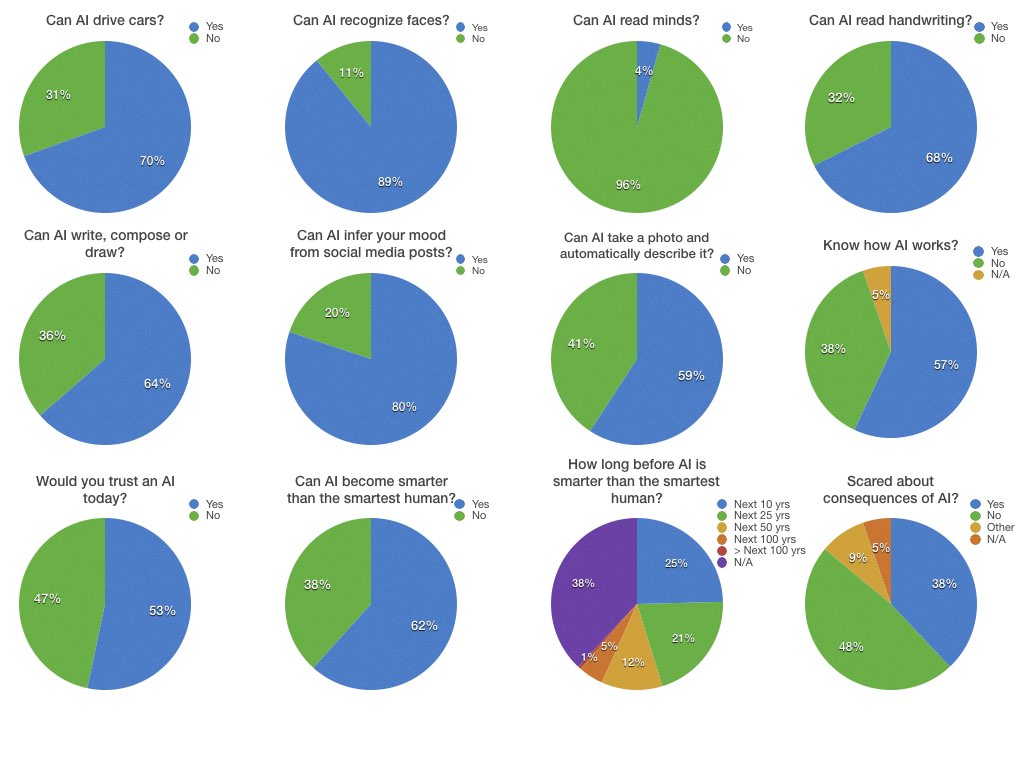}
 \vspace{-20pt}
 \caption{Perception of AI (across 321 subjects)}
 \vspace{-15pt}
\label{fig:ai_percep}
\end{figure*}

In Fig.~\ref{fig:pop_demog},~\ref{fig:tech_expo} and ~\ref{fig:ai_percep}, we break down the 321 subjects that completed the survey by their response to each question.

As part of the survey, subjects were also asked a few subjective questions about their opinions on present--day AI's capabilities. Specifically, they were asked to list tasks that they thought AI is capable of performing \emph{today} (see Fig.~\ref{fig:ai_now}), will be capable of \emph{in the next 3 years} (see Fig.~\ref{fig:ai_3_yrs}), and will be capable of \emph{in the next 10 years} (see Fig.~\ref{fig:ai_10_yrs}). We also asked how \emph{they} think AI works (see Fig.~\ref{fig:how_ai_works}). In Fig.~\ref{fig:ai_now}, ~\ref{fig:ai_3_yrs} and ~\ref{fig:ai_10_yrs}, we show word clouds corresponding to what subjects thought about the capabilities of AI. We also share some of those responses below.

\begin{compactenum}
\item Name three things that you think AI today can do.
\emph{Predict sports games; Detect specific types of cancer in images; Control house temp based on outside weather; translate; calculate probabilities; Predictive Analysis; AI can predict future events that happen like potential car accidents; lip reading; code; Facial recognition; Drive cars; Play Go; predict the weather; Hold a conversation; Be a personal assistant; Speech recognition; search the web quicker.}

\item Name three things that you think AI today can't yet do but will be able to do in 3 years.
\emph{Fly planes; Judge emotion in voices; Predict what I want for dinner; perform surgery; drive cars; manage larger amounts of information at a faster rate; think independently totally; play baseball; drive semi trucks; Be a caregiver; anticipate  a person's lying ability; read minds; Diagnose patients; improve robots to walk straight; Run websites; solve complex problems like climate change issues; program other ai; guess ages; form conclusions based on evidence; act on more complex commands; create art.}

\item Name three things that you think AI today can't yet do and will take a while ($>$ 10 years) before it can do it.
\emph{Imitate humans; be indistinguishable from humans; read minds; Have emotions; Develop feelings; make robots act like humans; truly learn and think; Replace humans; impersonate people; teach; be a human; full AI with personalities; Run governments; be able to match a human entirely; take over the world; Pass a Turing test; be a human like friend; intimacy; Recognize things like sarcasm and humor.}
\end{compactenum}

Interestingly, we observe a steady progression in subjects' expectations of AI's capabilities, as the time span increases. On a high-level reading through the responses, we notice that subjects believe that AI today can successfully perform tasks such as \emph{machine translation}, \emph{driving vehicles}, \emph{speech recognition}, \emph{analyzing information and drawing conclusions}, etc. (see Fig.~\ref{fig:ai_now}). It is likely that this is influenced by the subjects' exposure to or interaction with some form of AI in their day-to-day lives. When asked about what AI can do three years from now, most subjects suggested more sophisticated tasks such as \emph{inferring emotions from voice tone}, \emph{performing surgery}, and even \emph{dealing with climate change issues} (see Fig.~\ref{fig:ai_3_yrs}). However, the most interesting trends emerge while observing subjects' expectation of what AI can achieve in the next 10 years (see Fig.~\ref{fig:ai_10_yrs}). A major proportion of subjects believe that AI will gain the ability to \emph{understand and emulate human beings}, \emph{teach human beings}, \emph{develop feelings and emotions} and \emph{pass the Turing test}. 

We also observe how subjects think AI works (see Fig.~\ref{fig:how_ai_works}). Mostly, subjects believe that an AI agent today is a system with high computational capabilities that has been programmed to simulate intelligence  and perform certain tasks by exposing it to huge amounts of information, or, as one of subjects phrased it -- \emph{broadly AI recognizes patterns and creates optimal actions based on those patterns towards some predefined goals}. In summary, it appears that subjects have high expectations from AI, given enough time. While it is uncertain at this stage how many, or how soon, these feats will actually be achieved, we believe that building Theory of AI's mind skills will help humans generally become more active and effective collaborators in human--AI teams. 

As an interesting tidbit: Fig.~\ref{fig:ai_percep} shows what \% of subjects think certain tasks are ``solved''. 80-90\% of the subjects think AI today can recognize faces and infer your mood from social media posts. 65-70\% of subjects think AI today can recognize handwriting, be creative (write, compose, draw), or drive a car. They are more split on whether AI can describe an image in a sentence. However, most (96\%) agree that AI today cannot read our minds! Interestingly, 62\% of subjects think that AI can become smarter than the smartest human. \par


We now provide a full list of all questions in the survey.

\begin{compactenum}
\item How old are you?
	\begin{compactenum}
	\item Less than 20 years
    \item Between 20 and 40 years
    \item Between 40 and 60 years
    \item Greater than 60 years
	\end{compactenum}
\item What is your gender?
 	\begin{compactenum}
 	\item Male
    \item Female
    \item Other
 	\end{compactenum}
\item Where do you live?
 	\begin{compactenum}
 	\item Rural
 	\item Suburban
 	\item Urban
 	\end{compactenum}
\item Are you?
 	\begin{compactenum}
 	\item A student
 	\item Employed
 	\item Self-employed
 	\item Unemployed
 	\item Retired 
 	\item Other
 	\end{compactenum}
\item To which income group do you belong?
 	\begin{compactenum}
 	\item Less than 5000\$ per year
 	\item 5,000-10,000\$ per year
 	\item 10,000-25,000\$ per year
 	\item 25,000-60,000\$ per year 
 	\item 60,000-120,000\$ per year 
 	\item More than 120,000\$ per year
 	\end{compactenum}
\item What is your highest level of education?
 	\begin{compactenum}
 	\item No formal education
 	\item Middle School
 	\item High School
 	\item College (Bachelors)
 	\item Advanced Degree
 	\end{compactenum}
\item What was your major?
 	\begin{compactenum}
 	\item Computer Science / Computer Engineering
 	\item Engineering but not Computer Science
 	\item Mathematics / Physics
 	\item Philosophy
 	\item Biology / Physiology / Neurosciences
 	\item Psychology / Cognitive Sciences
 	\item Other Sciences
 	\item Liberal Arts
 	\item Other
 	\item None
 	\end{compactenum}
\item Do you know how to program / code?
 	\begin{compactenum}
 	\item Yes
 	\item No
 	\end{compactenum}
\item Does your full-time job involve:
 	\begin{compactenum}
 	\item No computers
 	\item Working with computers but no programming / coding?
 	\item Programming / Coding
 	\end{compactenum}
\item How many hours a day do you spend on your computer / laptop / smartphone?
 	\begin{compactenum}
 	\item Less than 1 hour
 	\item 1-5 hours
 	\item 5-10 hours
 	\item Above 10 hours
 	\end{compactenum} 
\item Do you know what Watson is in the context of Jeopardy?
 	\begin{compactenum}
 	\item Yes
 	\item No
 	\end{compactenum}
\item Have you ever used Siri, Alexa, or Google Now/Google Assistant?
 	\begin{compactenum}
 	\item Yes
 	\item No
 	\end{compactenum}
\item How often do you use Siri, Alexa, Google Now, Google Assistant, or something equivalent?
 	\begin{compactenum}
 	\item About once every few months
 	\item About once a month
 	\item About once a week
 	\item About 1-3 times a day
 	\item More than 3 times a day
 	\end{compactenum}
\item Have you heard of AlphaGo?
 	\begin{compactenum}
 	\item Yes
 	\item No
 	\end{compactenum}
\item Have you heard of Machine Learning?
 	\begin{compactenum}
 	\item Yes
 	\item No
 	\end{compactenum}
\item Have you heard of Deep Learning?
 	\begin{compactenum}
 	\item Yes
 	\item No
 	\end{compactenum}
\item When did you first hear of Artificial Intelligence (AI)?
	\begin{compactenum}
	\item I have not heard of AI
	\item More than 10 years ago
	\item 5-10 years ago
	\item 3-5 years ago
	\item 1-3 years ago
	\item In the last six months
	\item Last month
	\end{compactenum}
\item How did you learn about AI?
	\begin{compactenum}
	\item School / College
	\item Conversation with people
	\item Movies
	\item Newspapers
	\item Social media
	\item Internet
	\item TV
	\item Other
	\end{compactenum}
\item Do you think AI today can drive cars fully autonomously?
	\begin{compactenum}
	\item Yes
	\item No
	\end{compactenum}
\item Do you think AI today can automatically recognize faces in a photo?
	\begin{compactenum}
	\item Yes
	\item No
	\end{compactenum}
\item Do you think AI today can read your mind?
	\begin{compactenum}
	\item Yes 
	\item No
	\end{compactenum}
\item Do you think AI today can automatically read your handwriting?
	\begin{compactenum}
	\item Yes
	\item No
	\end{compactenum}
\item Do you think AI today can write poems, compose music, make paintings?
	\begin{compactenum}
	\item Yes
	\item No
	\end{compactenum}
\item Do you think AI today can read your Tweets, Facebook posts, etc. and figure out if you are having a good day or not?
	\begin{compactenum}
	\item Yes
	\item No
	\end{compactenum}
\item Do you think AI today can take a photo and automatically describe it in a sentence?
	\begin{compactenum}
	\item Yes
	\item No
	\end{compactenum}
\item Other than those mentioned above, name three things that you think AI today can do.
\item Other than those mentioned above, name three things that you think AI today can't yet do but will be able to do in 3 years.
\item Other than those mentioned above, name three things that you think AI today can't yet do and will take a while ($>$ 10 years) before it can do it.
\item Do you have a sense of how AI works?
	\begin{compactenum}
	\item Yes
	\item No
	\item If yes, describe in a sentence or two how AI works.
	\end{compactenum}
\item Would you trust an AI's decisions today?
	\begin{compactenum}
	\item Yes
	\item No
	\end{compactenum}
\item Do you think AI can ever become smarter than the smartest human?
	\begin{compactenum}
	\item Yes
	\item No
	\end{compactenum}
\item If yes, in how many years?
	\begin{compactenum}
	\item Within the next 10 years
	\item Within the next 25 years
	\item Within the next 50 years
	\item Within the next 100 years
	\item In more than 100 years
	\end{compactenum}
\item Are you scared about the consequences of AI?
	\begin{compactenum}
	\item Yes
	\item No
	\item Other
	\item If other, explain.
	\end{compactenum}
\end{compactenum}
\vspace{-5pt}

\end{document}